\documentclass[10pt,twocolumn,letterpaper]{article}

\usepackage{cvpr}              % To produce the CAMERA-READY version
\definecolor{cvprblue}{rgb}{0.21,0.49,0.74}
\usepackage[pagebackref,breaklinks,colorlinks,allcolors=cvprblue]{hyperref}

\def\paperID{} % *** Enter the Paper ID here
\def\confName{CVPR}
\def\confYear{2026}

\usepackage[accsupp]{axessibility}

\usepackage{booktabs}    % for \toprule, \midrule, \bottomrule
\usepackage{adjustbox}   % for \begin{adjustbox}
\usepackage{multirow}
\usepackage{graphicx}
\usepackage{amsmath}
\usepackage{amsfonts}
\usepackage[capitalize]{cleveref}
\usepackage{xcolor}
\usepackage{pifont}
\usepackage{array}
\usepackage{cuted}

\begin{document}

\title{Uni3R: Unified 3D Reconstruction and Semantic Understanding via \\Generalizable Gaussian Splatting from Unposed Multi-View Images}

%%%%%%%%% AUTHORS - PLEASE UPDATE
\newcommand\CoAuthorMark{\footnotemark[\arabic{footnote}]}
% \and

% \author{
% Xiangyu Sun$^{1}$\thanks{Equal contribution, intern at Horizon Robotics}\hspace{2.2mm}
% Haoyi Jiang$^{2}$\protect\CoAuthorMark\hspace{2.2mm}
% Liu Liu$^{4}$\thanks{Project leader}\hspace{2.2mm}
% Seungtae Nam$^{3}$\hspace{2.2mm}
% Gyeongjin Kang$^{1}$\hspace{2.2mm} \\
% Xinjie Wang$^{4}$\hspace{2.2mm}
% Wei Sui$^{5}$\hspace{2.2mm}
% Zhizhong Su$^{4}$\hspace{2.2mm}
% Wenyu Liu$^{2}$\hspace{2.2mm}
% Xinggang Wang$^{2}$\hspace{2.2mm}
% Eunbyung Park$^{3}$\thanks{Corresponding author}
% \vspace{2mm} \\
% $^1$Sungkyunkwan University\hspace{2.2mm}$^2$Huazhong University of Science \& Technology\hspace{2.2mm} \\
% $^3$Yonsei University\hspace{2.2mm}$^4$Horizon Robotics\hspace{2.2mm}$^5$D-Robotics
% \vspace{2mm} \\
% {\small \url{https://horizonrobotics.github.io/robot_lab/uni3R/}}
% }

\author{
Xiangyu Sun$^{1}$\thanks{Equal contribution. Intern at Horizon Robotics}\hspace{2.2mm}
Haoyi Jiang$^{2}$\protect\CoAuthorMark\thanks{Equal contribution. Intern at D-Robotics}\hspace{2.2mm}
Liu Liu$^{4}$\thanks{Project leader}\hspace{2.2mm}
Seungtae Nam$^{3}$\hspace{2.2mm}
Gyeongjin Kang$^{1}$\hspace{2.2mm} \\
Xinjie Wang$^{4}$\hspace{2.2mm}
Wei Sui$^{5}$\hspace{2.2mm}
Zhizhong Su$^{4}$\hspace{2.2mm}
Wenyu Liu$^{2}$\hspace{2.2mm}
Xinggang Wang$^{2}$\hspace{2.2mm}
Eunbyung Park$^{3}$\thanks{Corresponding author}
\vspace{2mm} \\
$^1$Sungkyunkwan University\hspace{2.2mm}$^2$Huazhong University of Science \& Technology\hspace{2.2mm} \\
$^3$Yonsei University\hspace{2.2mm}$^4$Horizon Robotics\hspace{2.2mm}$^5$D-Robotics
\vspace{2mm} \\
{\small \url{https://horizonrobotics.github.io/robot_lab/uni3R/}}
}

\maketitle

\begin{strip}
  \centering
  \vspace*{-20mm}
  \includegraphics[width=\textwidth]{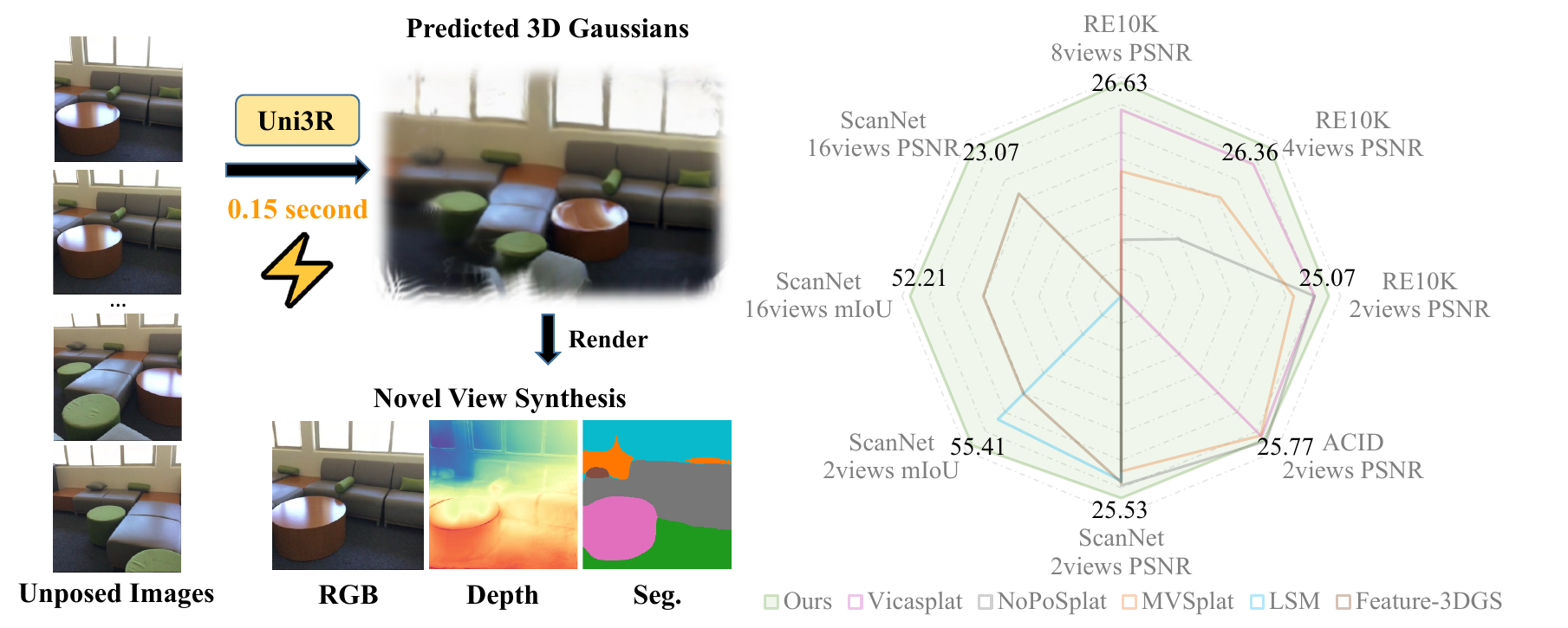}
  \vspace*{-6mm}
  \captionof{figure} {Uni3R takes unposed arbitrary multi-view images as input and produces a unified 3D Gaussian scene representation, enabling state-of-the-art performance in view synthesis, semantic segmentation, and depth estimation within a single forward pass.}
  \label{fig:pipeline}
\end{strip}

\begin{abstract}

Reconstructing and semantically interpreting 3D scenes from sparse 2D views remains a fundamental challenge in computer vision. Conventional methods often decouple semantic understanding from reconstruction or necessitate costly per-scene optimization, thereby restricting their scalability and generalizability. In this paper, we introduce \textbf{Uni3R}, a novel feed-forward framework that jointly reconstructs a unified 3D scene representation enriched with open-vocabulary semantics, directly from unposed multi-view images. Our approach leverages a Cross-View Transformer to robustly integrate information across arbitrary multi-view inputs, which then regresses a set of 3D Gaussian primitives endowed with semantic feature fields. This unified representation facilitates high-fidelity novel view synthesis, open-vocabulary 3D semantic segmentation, and depth prediction—all within a single, feed-forward pass.
Extensive experiments demonstrate that Uni3R sets a new state of the art across multiple benchmarks, including in-domain datasets such as RE10K and ScanNet, as well as the out-of-domain dataset Mip-NeRF360. This work represents a new paradigm toward generalizable and unified 3D scene reconstruction and understanding.
% The code is available at https://github.com/HorizonRobotics/Uni3R.
% Extensive experiments demonstrate that Uni3R establishes a new state-of-the-art across multiple benchmarks, including \textbf{25.07} PSNR on RE10K and \textbf{55.84} mIoU on ScanNet. Our work signifies a novel paradigm towards generalizable, unified 3D scene reconstruction and understanding. The code is available at https://github.com/HorizonRobotics/Uni3R.

% The ABSTRACT is to be in fully justified italicized text, at the top of the left-hand column, below the author and affiliation information.
% Use the word ``Abstract'' as the title, in 12-point Times, boldface type, centered relative to the column, initially capitalized.
% The abstract is to be in 10-point, single-spaced type.
% Leave two blank lines after the Abstract, then begin the main text.
% Look at previous \confName abstracts to get a feel for style and length.
\end{abstract}    
\vspace*{-6mm}
\section{Introduction}
\label{sec:intro}

The ability to perceive and interpret the 3D world from sparse images is a cornerstone of computer vision, holding profound implications for robotics, autonomous driving, and augmented reality. While significant progress has been made in 3D reconstruction, led by photorealistic methods such as Neural Radiance Fields (NeRF)~\cite{NeRF}, 3D Gaussian Splatting (3DGS)~\cite{3DGS}, their reliance on time-consuming, per-scene optimization critically limits their generalizability to novel scenes. In response, a prominent class of generalizable 3D reconstruction methods~\cite{pixelNeRF, pixelSplat, MVSplat, mvsgaussian, DepthSplat} has emerged, which learn geometric priors across diverse scenes to perform feed-forward 3D reconstruction in a feed-forward manner.

While promising, these methods typically focus exclusively on geometry and appearance, overlooking the semantic richness crucial for holistic scene understanding. Recent efforts, including LangSplat~\cite{LangSplat} and Feature-3DGS~\cite{Feature3DGS}, have incorporated semantic fields into 3D Gaussian Splatting, yet remain constrained by scene-specific optimization and lack scalability in real-world, zero-shot applications.
More recently, approaches such as LSM~\cite{LSM} and UniForward~\cite{UniForward} have aimed to unify semantic and radiance fields to jointly infer geometry, appearance, and semantics.
However, these methods are built upon DUSt3R~\cite{DUSt3R}, which is inherently designed for two-view inputs. Consequently, extending them to multi-view scenarios requires expensive pairwise feature matching across views, compromising efficiency and leading to inconsistent reconstructions due to the absence of global 3D context.

To address these limitations, we propose Uni3R, a novel, generalizable framework that synthesizes a unified 3D representation from arbitrary multi-view images for both high-fidelity rendering and dense, open-vocabulary semantic understanding.
Leveraging a Cross-View Transformer effectively fuses information across views and produces globally consistent representations, Uni3R predicts unified 3D Gaussian primitives enriched with open-vocabulary semantic features. These Gaussian representations can be seamlessly rendered in real-time to synthesize novel views, supervised solely with source images and bypassing the need for per-scene optimization. Simultaneously, the embedded semantic features enable zero-shot 3D semantic segmentation by querying the scene with arbitrary text prompts.

% To further enhance the geometric fidelity of reconstruction and training stability, we introduce a point-map-guided geometric loss that serves two key purposes. First, it improves structural consistency and leads to more accurate geometry, as reflected by lower depth errors (e.g., AbsRel). Second, it stabilizes training by preventing the model from getting trapped in local minima when predicting the freedom 3D point distribution. Specifically, we employ a frozen VGGT~\cite{VGGT} to generate dense point maps along with associated confidence scores, which act as soft priors to guide the spatial distribution of the 3D Gaussians.

To further enhance both geometric fidelity and training stability, we introduce a point-map-guided geometric loss that serves two key purposes. First, it enforces structural consistency and improves geometric accuracy, as evidenced by lower depth errors (e.g., AbsRel). Second, it stabilizes training by preventing the model from getting trapped in local minima when predicting the freedom 3D point distribution. Specifically, we employ a frozen VGGT~\cite{VGGT} to generate dense point maps with associated confidence scores, which act as soft geometric priors to guide the spatial distribution of the 3D Gaussians.

Our contributions are summarized as follows:

% We introduce Uni3R, a novel feed-forward architecture that unifies 3D reconstruction and semantic understanding. It predicts a set of Gaussian primitives with jointly integrated geometry, appearance, and open-vocabulary semantics in a single pass, eliminating the need for per-scene optimization.

% We demonstrate that a powerful geometry foundation model can be effectively extended beyond geometric estimation to support both photometric reconstruction and 3D scene understanding. Its cross-frame attention mechanism enables robust feature fusion to produce globally consistent scene representations from an arbitrary number of input views, while its predicted point maps provide potent geometric guidance.

% Uni3R achieves state-of-the-art performance across multiple tasks, including novel view synthesis, open-vocabulary 3D semantic segmentation, and depth prediction on the challenging RE10K~\cite{RE10K} and ScanNet~\cite{ScanNet} datasets, underscoring its superior generalization and versatility.

\begin{itemize}
\item We introduce Uni3R, a novel feed-forward architecture that unifies 3D reconstruction and semantic understanding. It predicts a set of Gaussian primitives with jointly integrated geometry, appearance, and open-vocabulary semantics in a single pass, eliminating the need for per-scene optimization.

\item We demonstrate that a powerful geometry foundation model can be effectively extended beyond geometric estimation to support both photometric reconstruction and 3D scene understanding. Its cross-frame attention mechanism enables robust feature fusion to produce globally consistent scene representations from an arbitrary number of input views, while its predicted point maps provide potent geometric guidance.

\item Uni3R achieves state-of-the-art performance across multiple tasks, including novel view synthesis, open-vocabulary 3D semantic segmentation, and depth prediction on the challenging RE10K~\cite{RE10K} and ScanNet~\cite{ScanNet} datasets, underscoring its superior generalization and versatility.
\end{itemize}

\section{Related Work}
\label{sec:related_work}

\subsection{Differentiable Neural Representations}
Traditional 3D reconstruction methods, such as Structure-from-Motion (SfM)~\cite{SfM} and Multi-View Stereo (MVS)~\cite{MVS} decompose the process into sequential steps, including feature matching, camera pose estimation, and geometric reconstruction. While effective, these multi-stage processes can be fragile and prone to error accumulation. The advent of Neural Radiance Fields (NeRF)~\cite{NeRF} revolutionized novel view synthesis by introducing an end-to-end, differentiable approach that models a scene as a continuous function mapping 5D coordinates to color and volumetric density. More recently, 3D Gaussian Splatting (3DGS)~\cite{3DGS, sun2024f3dgs, hu2025ccgs, lee2024compact3dgs} has emerged as a compelling alternative, representing scenes explicitly with a set of 3D Gaussian primitives. Leveraging a highly efficient differentiable rasterizer, 3DGS supports real-time rendering speeds while maintaining exceptional rendering quality. However, canonical 3DGS relies on point clouds from SfM for initialization and rectified camera poses. Our work builds upon the 3DGS formulation but removes reliance on external tools like COLMAP~\cite{SfM}. By predicting Gaussians in an end-to-end pose-free manner, we enable scalable 3D reconstruction and scene understanding.

\begin{figure*}[t]
  \centering
  \includegraphics[width=\textwidth]{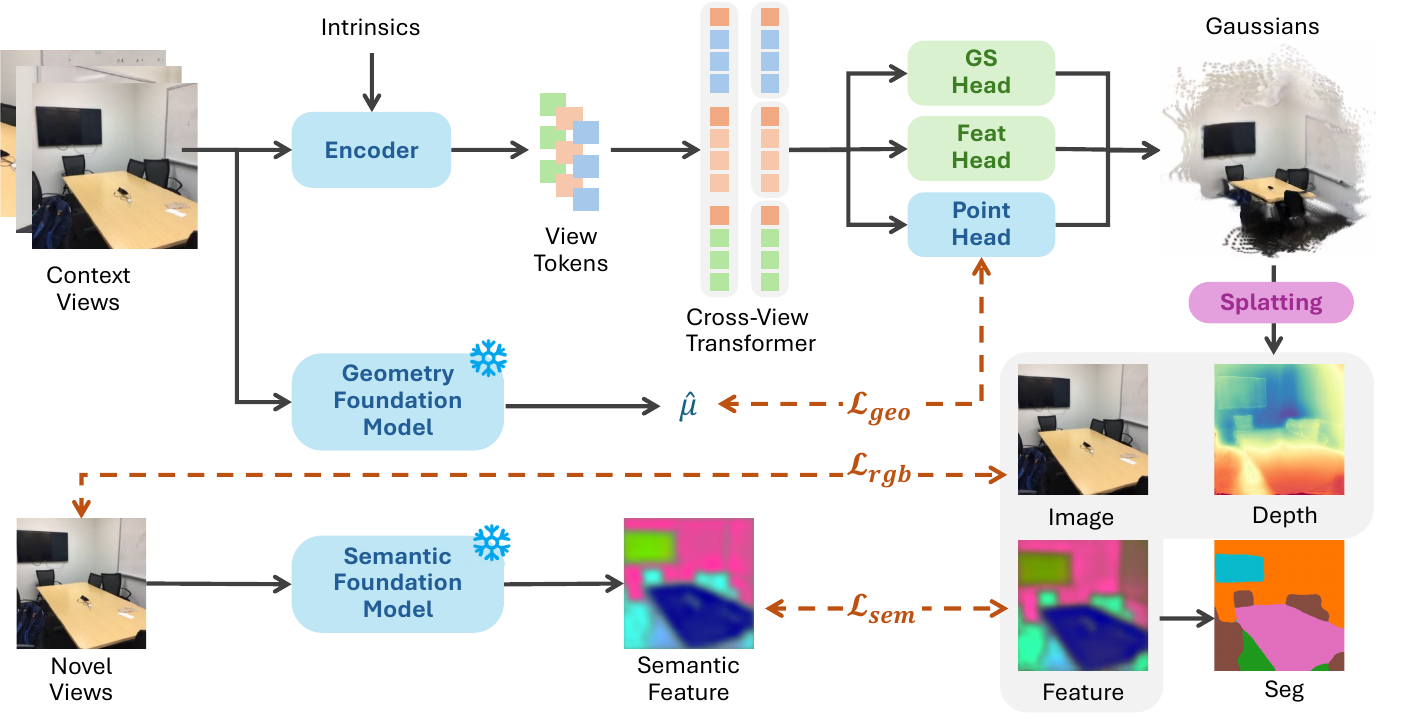}
  \caption{Architectural overview of the Uni3R pipeline. Uni3R predicts a set of Gaussian primitives with jointly integrated geometry, appearance, and open-vocabulary semantics in a single pass, eliminating the need for per-scene optimization.}
  \vspace{-5mm}
  \label{fig:framework}
\end{figure*}

\subsection{Feed-forward 3D Reconstruction}
The substantial computational cost of per-scene optimization has motivated the development of 3D feed-forward models. PixelNeRF~\cite{pixelNeRF} and MVSNeRF~\cite{MVSNeRF} learn scene priors across a large number of training scenes, enabling them to predict radiance fields for novel scenes from only a few input views. Following the success of 3DGS, pixelSplat~\cite{pixelSplat}, MVSplat~\cite{MVSplat}, Generative Densification~\cite{GD}, iLRM~\cite{ilrm3d} and DepthSplat~\cite{DepthSplat} adapt this generalizable paradigm to predict 3D Gaussian parameters directly. However, these approaches typically necessitate known camera poses to guide the reconstruction. To eliminate this constraint, MASt3R~\cite{MASt3R} and DUSt3R~\cite{DUSt3R} demonstrate the feasibility of predicting pixel-aligned 3D point clouds directly from image pairs without explicit pose information. Building on these advances, Splatt3R~\cite{Splatt3R} and NoPoSplat~\cite{NoPoSplat} further advance this pose-free paradigm by predicting 3D Gaussian primitives directly from image pairs. Despite their progress, models based on the DUSt3R architecture still require sufficient overlap between image pairs and struggle to integrate globally consistent information, leading to fragmented reconstructions.
Uni3R overcomes these limitations by employing a Cross-View Transformer, inspired by VGGT~\cite{VGGT}, to interpret and fuse information from an arbitrary number of views. Based on the globally consistent 3D geometric features, we develop a multi-view, pose-free feed-forward reconstruction model. Our method supports not only image pairs but also extended sequences or video clips, predicting 3D Gaussian primitives in a single forward pass to achieve high-quality, globally coherent 3D reconstruction without requiring camera poses.

\subsection{Open-Vocabulary Segmentation in 3DGS}
Integrating semantics into 3D reconstructions is crucial for higher-level scene understanding tasks. Early methods for 3D semantic segmentation required dense, 3D ground-truth labels, which are scarce and laborious to acquire. The advent of powerful 2D vision-language models like CLIP~\cite{CLIP, maskadapter} has spurred the development of open-vocabulary methods that lift 2D understanding into 3D. LERF~\cite{LERF} distills 2D CLIP features into 3D radiance fields. Capitalizing on the rendering efficiency of 3D Gaussian Splatting, several methods~\cite{LangSplat, Feature3DGS, panogs, supergseg} have extended Gaussian representations with semantic features. Nonetheless, these methods still rely on per-scene optimization, making them unsuitable for real-time applications in novel environments. While generalizable approaches like LSM~\cite{LSM} and GSemSplat~\cite{gsemsplat} have been proposed, they are typically constrained to two-view inputs, restricting their scalability and robustness in complex scenes. In a related vein, GaussTR~\cite{GaussTR} explores generalizable Gaussian-based segmentation in the context of occupancy prediction.
In contrast, Uni3R integrates open-vocabulary understanding into a generalizable, multi-view framework, producing globally consistent 3D representation embedded with expressive semantics without requiring any 3D semantic labels.

\section{Method}
\label{sec:method}

This section details our methodology, beginning with the Feed-Forward 3D Gaussian Model in \cref{sec:ffgs}. We then describe how to endow Gaussians with semantics in \cref{sec:rwoc}, and conclude with the specifics of the training losses in \cref{sec:losses}, including photometric loss, semantic loss and geometry loss.

\subsection{Feed-Forward Gaussian Splatting}
\label{sec:ffgs}

\subsubsection{Intrinsic Embedding}

To resolve the inherent scale ambiguity in monocular reconstruction caused by unknown focal lengths, we incorporate an intrinsic embedding to provide essential geometric cues. Following NoPoSplat~\cite{NoPoSplat}, we encode each camera's focal length and principal point with a linear projection. The resulting intrinsic embedding is concatenated channel-wise with the corresponding image before patch tokenization, allowing the network to reason about the geometry-aware information.

\subsubsection{Cross-View Transformer Encoder}

Uni3R employs a Cross-View Transformer Encoder, following VGGT, to extract and fuse features from all input images into a consistent, view-agnostic latent representation. Each input view $I^{(i)}$, augmented with its intrinsic embedding, is first processed by a pre-trained Vision Transformer, DINOv2~\cite{DINOv2}, to extract a sequence of patch-level feature tokens. To support arbitrary multi-view inputs while maintaining permutation equivariance, a learnable camera token is appended to each view's token sequence. The Cross-View Transformer Encoder consists of a series of Transformer blocks that alternate between intra-frame and cross-frame attention. Intra-frame self-attention operates within each view's token set, refining the per-view features with local context. Subsequently, cross-frame global attention aggregates tokens from all views to establish correspondences and reason about the global 3D geometry. The output latent tokens from the encoder encapsulate a holistic and globally consistent understanding of the 3D scene.

\subsubsection{Decoding Gaussian Parameters}

The fused latent representations are decoded into a dense set of 3D Gaussian primitives with a Dense Prediction Transformer (DPT)~\cite{DPT} followed by dedicated prediction heads for different Gaussian parameters. DPT progressively refines coarse patch-level features with fine-grained local details from intermediate layers, yielding a dense per-pixel feature map.

Subsequently, we predict the properties of a set of pixel-aligned 3D Gaussians with separate MLP heads. Each primitive is parameterized by:
\begin{equation}
    G_j = \{ \mu_j, \alpha_j, c_j, s_j, r_j, f_j^{\text{sem}} \},
\end{equation}
where $\mu_j \in \mathbb{R}^3$ denotes the 3D center point, $s_j \in \mathbb{R}^3$ is the scale, $r_j \in \mathbb{R}^4$ is the rotation quaternion, $\alpha_j \in [0, 1]$ is the opacity, $c_j \in \mathbb{R}^3$ is the color, and $f_j^\text{sem} \in \mathbb{R}^d$ is a high-dimensional semantic feature vector.

The point head is initialized from pre-trained VGGT weights and is further fine-tuned with rendering-based supervision to align with real-world metric scales. Distinct activation functions are applied to the predicted parameters to constrain them to their valid ranges:
\begin{align}
    \alpha_j &= \sigma(f_j^\alpha), \\
    s_j &= \exp(f_j^s) \cdot d_{\text{median}}, \\
    r_j &= \text{normalize}(f_j^r),
\end{align}
where $\sigma(\cdot)$ represents the sigmoid activation function, and $f_j^\alpha$, $f_j^s$, and $f_j^r$ are the latents for opacity, scale, and rotation, respectively. The term $d_\text{median}$ is the median depth value computed from the predicted 3D positions, which helps to normalize the scale.

\subsection{Rendering with Open-Vocabulary Semantics}
\label{sec:rwoc}

Once predicted, the set of Gaussians is rendered into novel views using the differentiable 3D Gaussian rasterizer, extended with semantic feature fields. The Gaussian function is described by:
\begin{equation}
    G_j(x) = e^{-\frac{1}{2}x^\top \Sigma_j^{-1}x},
\end{equation}
where the covariance matrix $\Sigma_j$ is constructed from the scale $s_j$ and rotation $r_j$. The rendered color $\hat{I}$ and feature $\hat{F}$ at each pixel are computed by alpha-blending the properties of all sorted Gaussians that overlap it, taking $\hat{F}$ as an example:
\begin{equation}
    \hat{F} = \sum_{i} \hat{f}_i^\text{sem} \alpha_i \prod_{j=1}^{i-1} (1 - \alpha_j),
\end{equation}
where $\hat{f}_j^\text{sem}$ is compressed from $f_j^\text{sem}$ by an autoencoder to mitigate the high memory cost of rendering high-dimensional semantic features.
\begin{align}
    \hat{f}_j^\text{sem} &= \mathcal{F}_{\text{enc}}(f_j^\text{sem}), \\
    \hat{F}' &= \mathcal{F}_{\text{dec}}(\hat{F}),
\end{align}
where $\mathcal{F}_{\text{enc}}$ and $\mathcal{F}_{\text{dec}}$ are the encoder and decoder, respectively. The autoencoder is trained end-to-end to align the rendered features with CLIP-based image features, enabling efficient open-vocabulary semantic reasoning.

During inference, semantic segmentation is performed by computing the cosine similarity between the pixel-wise semantic features and a set of text-derived prototypes. Given a set of text prompts for desired categories (\textit{e.g.}, ``wall," ``chair," ``sofa"), CLIP text encoder generates corresponding feature prototypes $f^\text{txt} \in \mathbb{R}^{N_C \times C}$, where $N_C$ is the number of categories. The semantic logits $S$ is then computed by cosine similarity:
\begin{equation}
S_p = \text{softmax}(f^\text{txt} \cdot \hat{F}').
\end{equation}

% \begin{figure*}[!t]
%   \centering
%   % \vspace*{-6mm}
%   \includegraphics[width=\textwidth]{figures/visualization.pdf}
%   \caption{\textbf{Qualitative Comparison of Novel-View Segmentation on ScanNet.}}
%   \label{fig:scannet_sem}
% \end{figure*}

\begin{figure*}[t]
  \centering
  \includegraphics[width=\textwidth]{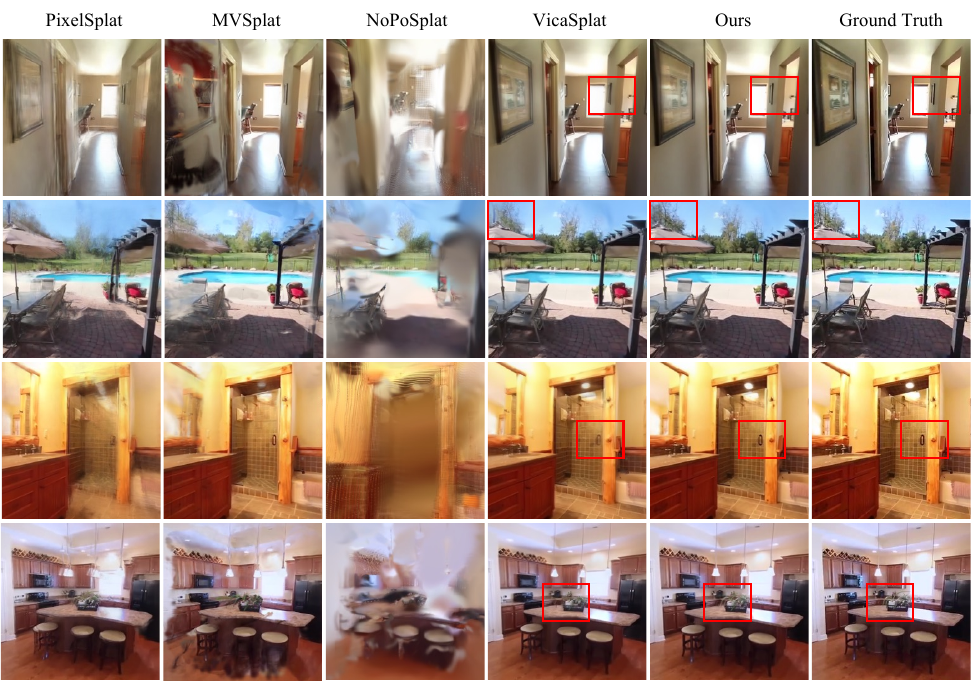}
  \caption{\textbf{Qualitative comparison of novel view synthesis on RealEstate10k test set with 8 input images.}}
  \vspace{-6mm}
  \label{fig:re10k_sequential}
\end{figure*}
                                                    
\subsection{Training Objectives}
\label{sec:losses}

\paragraph{Photometric Loss ($\mathcal{L}_{\text{rgb}}$).} To ensure that rendered images match the input views, we combines a pixel-wise L1 loss and the LPIPS metric~\cite{LPIPS}:
\begin{equation}
\mathcal{L}_{\text{rgb}} = \sum_{i=1}^N \left( ||\tilde{I}^{(i)} - \hat{I}^{(i)}||_1 + \lambda_{\text{LPIPS}}\text{LPIPS}(\tilde{I}^{(i)}, \hat{I}^{(i)}) \right), 
\end{equation}
where $\tilde{I}^{(i)}$ and $\hat{I}^{(i)}$ denotes the ground-truth image and the rendered image from the $i$-th camera viewpoint, respectively, and $\lambda_{\text{LPIPS}}$ is set to 0.05.

\begin{table*}[t]
    \centering
    \footnotesize
    % \vspace*{-6mm}
    \begin{adjustbox}{max width=\textwidth}
    \setlength{\tabcolsep}{0.08cm}
    \renewcommand{\arraystretch}{1.0}
    \begin{tabular}{l| cc | cc | cc | cc | ccc}
        \toprule
        & \multicolumn{2}{c}{Recon. Time$\downarrow$} & \multicolumn{4}{c}{Source View} & \multicolumn{5}{c}{Target View} \\
        \cmidrule(lr){2-3} \cmidrule(lr){4-7} \cmidrule(lr){8-12}
        Method & SfM & Per-Scene & mIoU$\uparrow$ & Acc.$\uparrow$ & rel$\downarrow$ & $\tau\uparrow$ & mIoU$\uparrow$ & Acc.$\uparrow$ & PSNR$\uparrow$ & SSIM$\uparrow$ & LPIPS$\downarrow$ \\
        \midrule
        LSeg & N/A & N/A & 0.4701 & 0.7891 & - & - & 0.4819 & 0.7927 & - & - & - \\
        NeRF-DFF & 20.52s & 1min & 0.4540 & 0.7173 & 27.68 & 9.61 & 0.4037 & 0.6755 & 19.86 & 0.6650 & 0.3629 \\
        Feature-3DGS & 20.52s & 18mins & 0.4453 & 0.7276 & 12.95 & 21.07 & 0.4223 & 0.7174 & 24.49 & 0.8132 & 0.2293  \\
        \midrule
        PixelSplat &  \multicolumn{2}{c|}{0.064s} & - & - & - & - & - & - & 24.89 & 0.8392 & 0.1641 \\
        LSM* & \multicolumn{2}{c|}{0.108s} & 0.5034 & 0.7740 & \textbf{3.38} & \textbf{67.77} & 0.5078 & 0.7686 & 24.39 & 0.8072 & 0.2506 \\
        AnySplat & \multicolumn{2}{c|}{-} & - & - & 6.35 & 47.57 & - & - & 22.08 & 0.8118 & 0.2480 \\
        Ours & \multicolumn{2}{c|}{0.162s} & \textbf{0.5403} & \textbf{0.8255} & 3.87  & 61.37  & \textbf{0.5584} & \textbf{0.8268} & \textbf{25.53} & \textbf{0.8727} & \textbf{0.1380}  \\
        % Ours w/o pm loss & \multicolumn{2}{c|}{0.159s} & 0.5388 & 0.8218 & 5.81 & 47.99 & \textbf{0.5541} & 0.8221 & \textbf{25.53} & 0.8719 & 0.1411 \\
        % Ours w/ pm loss 0.005 & \multicolumn{2}{c|}{0.162s} & \textbf{0.5403} & \textbf{0.8255} & 3.92  & 61.24  & 0.5484 & \textbf{0.8265} & \textbf{25.53} & \textbf{0.8727} & \textbf{0.1380}  \\
        % Ours w/ pm loss 0.01 & \multicolumn{2}{c|}{0.152s} & 0.5349 & 0.8261 & 3.73 & 63.25 & 0.5522 & 0.8268 & 25.39 & 0.8707 & 0.1389 \\
        \bottomrule
    \end{tabular}
    \end{adjustbox}

    % \caption{\textbf{Quantitative Comparison in 3D Tasks. * indicates that LSM uses ground truth point cloud to do supervision while Uni3R is self-superivision. } We evaluate our model across three core 3D vision tasks: novel view synthesis, depth estimation, and open-vocabulary semantic segmentation. For segmentation, we report results based on our evaluation protocol using LSeg.}
    \caption{\textbf{Quantitative Comparison on ScanNet.} We evaluate performance on novel view synthesis, depth estimation, and open-vocabulary semantic segmentation. \textbf{(*) Unlike LSM, Uni3R is trained without any 3D annotations.}
}

\label{tab:scannet_semantic}
\end{table*}

\paragraph{Semantic Loss ($\mathcal{L}_{\text{sem}}$).} To endow the Gaussians with open-vocabulary capabilities, we distill knowledge from a frozen, pre-trained 2D vision-language model, LSeg~\cite{LSeg}. We extract feature maps $\tilde{F}^{(i)}$ from each input image using the LSeg image encoder. We then enforce alignment between the rendered semantic feature map $\hat{F}^{(i)'}$ and the 2D CLIP-based features using a cosine similarity loss:
\begin{equation}
\mathcal{L}_{\text{sem}} = \sum_{i=1}^N \left( 1 - \frac{\tilde{F}^{(i)} \cdot \hat{F}^{(i)'}}{||\tilde{F}^{(i)}|| \cdot ||\hat{F}^{(i)'}||} \right).
\end{equation}
This loss lifts rich 2D semantics into the 3D domain, enabling zero-shot semantic understanding without requiring explicit 3D annotations.

\paragraph{Geometry Loss ($\mathcal{L}_{\text{geo}}$).}

To enhance geometric consistency and training stability, we adopt a point-map regularization strategy inspired by PM-Loss~\cite{Pmloss}. This regularization simultaneously improves structural accuracy—particularly around object boundaries—and mitigates collapse during optimization. Under RGB-only supervision, the model lacks explicit geometric constraints on the predicted point cloud, often leading to local minima and unstable convergence. The introduced point-map constraint provides a strong geometric prior that guides the 3D Gaussian distribution toward both structurally consistent and stable reconstruction.
% To enhance the geometric consistency, particularly around object boundaries, we adopt a point-map regularization strategy inspired by PM-Loss~\cite{Pmloss}. This regularization not only improves structural accuracy but also stabilizes training, preventing collapse during optimization. 
Specifically, we leverage a frozen VGGT~\cite{VGGT} to generate a dense point map $\hat{\mu}^{(i)} \in \mathbb{R}^{3 \times H \times W}$ to guide geometric supervision. 
Given that the predictions from VGGT are not uniformly reliable, especially in challenging regions such as reflective surfaces or areas with heavy occlusion, we introduce a confidence-based masking strategy. We extract the confidence map $C^{(i)}$ from VGGT and construct a binary geometry mask, $M^{(i)} \in \{0,1\}^{H\times W}$ by selecting the top-$k$ most confident pixels (set as $90\%$ in our experiments). The predicted point maps $\mu^{(i)}$ from Uni3R are then aligned with $\hat{\mu}^{(i)}$ via the Umeyama algorithm~\cite{Umeyama}. Given the masked aligned point clouds $X_{U}^{(i)}=\mu^{(i)}\odot M^{(i)}$ and $X_{V}^{(i)}=\hat{\mu}^{(i)}\odot M^{(i)}$, where $\odot$ denotes the element-wise product, a single-directional Chamfer distance is computed. The loss is formulated as:
\begin{equation}
\mathcal{L}_{\text{geo}} = \sum_{i=1}^{N}\frac{1}{N_{pts}^{(i)}}\sum_{x \in X_{U}^{(i)}}\min_{x' \in X_{V}^{(i)}}||x-x'||_2^2
\end{equation}
$N_{pts}^{(i)}$ is the total number of points. The final training objective is a weighted sum of the individual losses:
\begin{equation}
    \mathcal{L}_{\text{total}} = \mathcal{L}_{\text{rgb}} + \lambda_{\text{sem}}\mathcal{L}_{\text{sem}} +   \lambda_{\text{geo}}\mathcal{L}_{\text{geo}},
\end{equation}
where the balancing hyperparameters $\lambda_{\text{sem}}$ and $\lambda_{\text{geo}}$ are set to 0.02 and 0.005, respectively.
\section{Experiments}
\label{sec:experiments}

\subsection{Experimental Setup}

\newcommand{\cmark}{\raisebox{0.2ex}{\ding{51}}}
\newcommand{\xmark}{\raisebox{0.2ex}{\ding{55}}}

\begin{table*}[!t]
    \centering
    \begin{adjustbox}{max width=\linewidth}
    \setlength{\tabcolsep}{0.08cm}
    \renewcommand{\arraystretch}{1.0}
    \begin{tabular}{l|c| c | cc | cc | cc | c | cc | cc | cc}
        \toprule
        && \multicolumn{7}{c}{8 views} & \multicolumn{7}{c}{16 views} \\ 
        \cmidrule(lr){3-9} \cmidrule(lr){10-16}
        Method & Feed-Forward & Time$\downarrow$ & PSNR$\uparrow$ & SSIM$\uparrow$ & mIoU$\uparrow$ & Acc.$\uparrow$ & rel$\downarrow$ & $\tau\uparrow$ & Time$\downarrow$ & PSNR$\uparrow$ & SSIM$\uparrow$ & mIoU$\uparrow$ & Acc.$\uparrow$ & rel$\downarrow$ & $\tau\uparrow$ \\
        \midrule
        Feature-3DGS & \xmark & $\approx40min$ & 18.17 & 0.674 & 0.195 & 0.724 & 17.28 & 13.31 & $\approx60min$ & 17.09 & 0.649 & 0.198  & 0.672 & 23.71 & 10.57 \\
        % DFFs &&&  &  &  &&&  &  & && \\
        \textbf{Ours} & \cmark & \textbf{0.359s} & \textbf{24.71} & \textbf{0.851} & \textbf{0.554} & \textbf{0.807} & \textbf{4.25} & \textbf{57.41} & \textbf{0.636s} & \textbf{23.32} & \textbf{0.816} & \textbf{0.558} & \textbf{0.835} & \textbf{3.97} & \textbf{60.48} \\
        \bottomrule
    \end{tabular}
    \end{adjustbox}
    \caption{\textbf{Comparison with Per-Scene Optimized Methods.} Time corresponds to the average reconstruction time per scene.}

\label{tab:scannet_semantic_multiviews}
\end{table*}

% \begin{figure*}[t]
%   \centering
%   \includegraphics[width=\textwidth]{figures/re10k_backup.pdf}
%   \caption{\textbf{Qualitative comparison of novel view synthesis on RealEstate10k test set with 8 input images.}}
%   \label{fig:re10k_sequential}
% \end{figure*}

\begin{figure*}[!t]
  \centering
  % \vspace*{-6mm}
  \includegraphics[width=\textwidth]{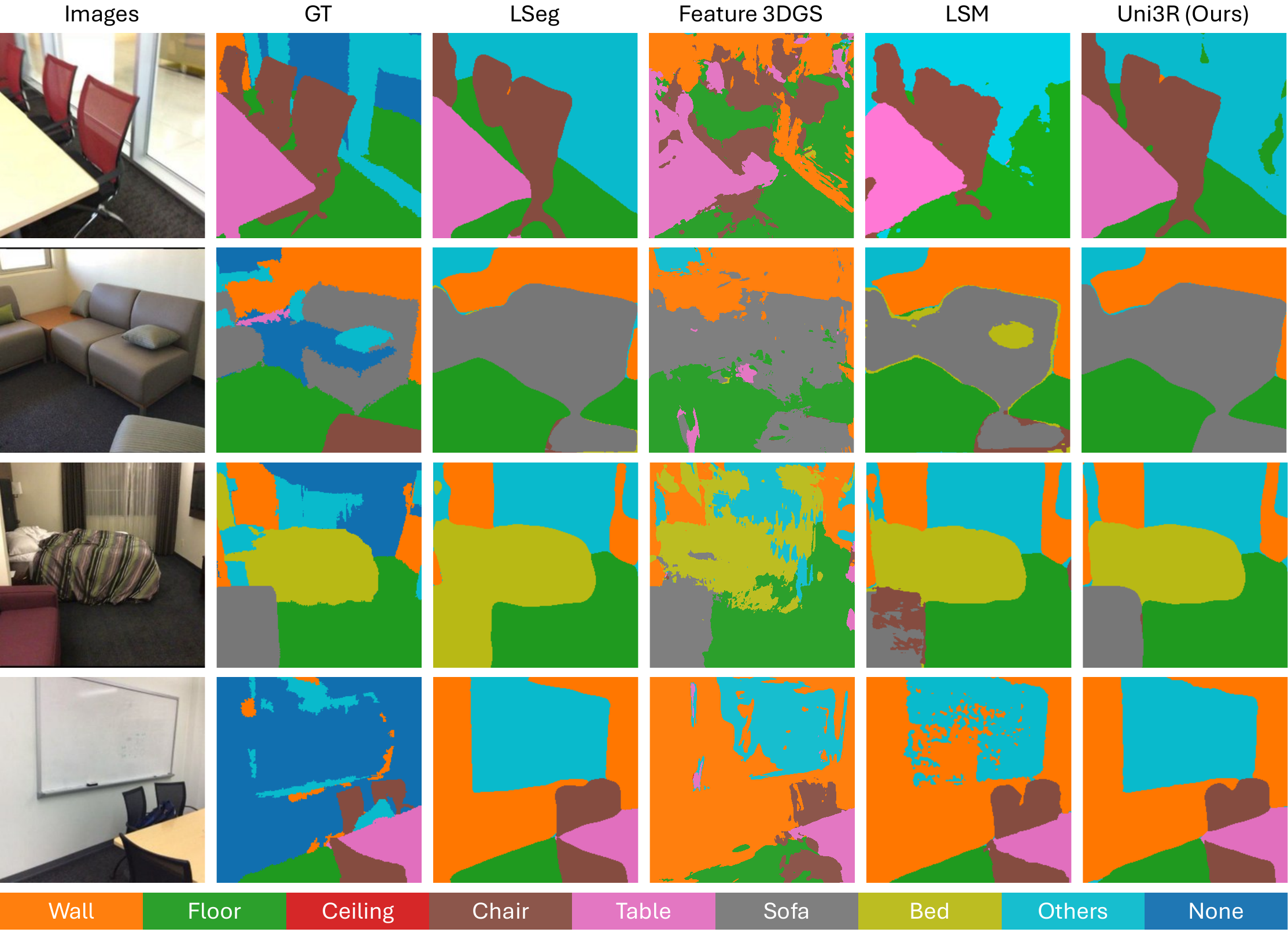}
  \caption{\textbf{Qualitative Comparison of Novel-View Segmentation on ScanNet.}}
  \vspace{-5mm}
  \label{fig:scannet_sem}
\end{figure*}

\paragraph{Dataset}

For evaluating both 3D scene and semantic field reconstruction, our model is trained on a combined dataset of ScanNet++~\cite{ScanNet++} and ScanNet~\cite{ScanNet}, totaling 1,565 scenes. We evaluate on 40 unseen ScanNet scenes, and further examine the model’s zero-shot generalization on the Mip-NeRF360~\cite{barron2022mip} dataset.

Furthermore, to assess rendering quality, we train our model on the RealEstate10K~\cite{RE10K} and ACID~\cite{ACID} datasets. To evaluate cross-domain generalization, we test our method on the DTU~\cite{DTU} and ScanNet++~\cite{ScanNet++} benchmarks (see the supplementary material for more details).

% Furthermore, to evaluate rendering quality, we train our model on the RealEstate10K~\cite{RE10K} and ACID~\cite{ACID} datasets. For cross-domain generalization ability, we evaluate our method on the DTU~\cite{DTU} and ScanNet++~\cite{ScanNet++} benchmarks (please refer to the supplementary material).

\paragraph{Implementation Details}
We use DINOv2~\cite{DINOv2} as the image encoder, with a patch size of 16, and set the Cross-View Transformer layers as $L=24$. We initialize the encoder and decoder with the weights from the pretrained VGGT~\cite{VGGT}, while the remaining intrinsic layer and Gaussian head are randomly initialized. For a fair comparison with the baseline models, we report all quantitative results under $256\times256$. Our model is implemented using PyTorch~\cite{PyTorch}. 
% with a maximum image input length of $N = 16$. 
All experiments are conducted on 8 $\times$ A100 GPUs, taking approximately 22 hours for the training of 2 views, with a batch size of 2. 
Please refer to the supplementary for more details.
% To evaluate both 3D rendering and semantic performance, we perform novel view synthesis and segmentation with peak signal-to-noise ratio (PSNR), structural similarity index (SSIM)~\cite{SSIM}, perceptual distance (LPIPS)~\cite{LPIPS}, mean intersection over union (mIoU) and mean pixel accuracy (mAcc) as metrics.

\begin{table}[!t]
    \centering
    \begin{adjustbox}{max width=\linewidth}
    \setlength{\tabcolsep}{0.08cm}
    \renewcommand{\arraystretch}{1.0}
    \begin{tabular}{llccc | ccc}
        \toprule
        & & \multicolumn{3}{c}{RE10k} & \multicolumn{3}{c}{ACID} \\ 
        \cmidrule(lr){3-5} \cmidrule(lr){6-8}
        & Method & PSNR$\uparrow$ & SSIM$\uparrow$ & LPIPS$\downarrow$ & PSNR$\uparrow$ & SSIM$\uparrow$ & LPIPS$\downarrow$ \\
        \midrule
        \multirow{2}{*}{\shortstack[l]{\emph{Pose-} \\ \emph{Required}}} 
        & PixelSplat & 23.361 & 0.803 & 0.186 & 25.684 & \underline{0.778} & 0.194 \\
        & MVSplat & 23.430 & 0.805 & 0.179 & 25.335 & 0.772 & 0.195 \\
        \midrule
        \multirow{4}{*}{\shortstack[l]{\emph{Pose-} \\ \emph{Free}}} 
        % & Splatt3R & - & - & - & - & - & - \\
        & NoPoSplat & 25.036 & \textbf{0.838} & 0.162 & \textbf{25.961} & \textbf{0.781} & \textbf{0.189} \\
        & CoPoNeRF & 18.938 & 0.619 & 0.388 & 20.950 & 0.606 & 0.406 \\
        & VicaSplat & \underline{25.038} & 0.834 & \underline{0.161} & 25.439 & 0.757 & 0.201 \\
        & \textbf{Ours} & \textbf{25.074} & \underline{0.837} & \textbf{0.158} & \underline{25.766} & \underline{0.778} & \underline{0.192} \\
        \bottomrule
    \end{tabular}
    \end{adjustbox}
    \caption{\textbf{Quantitative comparisons of novel view synthesis on the RE10k~\cite{RE10K} and ACID~\cite{ACID} dataset under 2-views setup.}}
\label{tab:re10k_acid_2view}
\end{table}

\begin{table*}[t]
    \centering
    \footnotesize
    \begin{adjustbox}{max width=\textwidth}
    \setlength{\tabcolsep}{0.08cm}
    \renewcommand{\arraystretch}{1.0}
    \begin{tabular}{llccc ccc | ccc ccc}
        \toprule
        & & \multicolumn{3}{c}{4 views (RE10k)} & \multicolumn{3}{c}{8 views (RE10k)} & \multicolumn{3}{c}{4 views (ScanNet)} & \multicolumn{3}{c}{8 views (ScanNet)} \\
        \cmidrule(lr){3-5} \cmidrule(lr){6-8} \cmidrule(lr){9-11} \cmidrule(lr){12-14}
        & Method & PSNR$\uparrow$ & SSIM$\uparrow$ & LPIPS$\downarrow$ & PSNR$\uparrow$ & SSIM$\uparrow$ & LPIPS$\downarrow$ & PSNR$\uparrow$ & SSIM$\uparrow$ & LPIPS$\downarrow$ & PSNR$\uparrow$ & SSIM$\uparrow$ & LPIPS$\downarrow$ \\
        \midrule
        \multirow{4}{*}[0.7em]{\centering \shortstack[l]{\emph{Pose-} \\ \emph{Required}}} 
        % & pixelNeRF & 18.417 & 0.601 & 0.526 & 19.930 & 0.632 & 0.480 & 20.869 & 0.639 & 0.458 & 19.824 & 0.626 & 0.485 \\
        & PixelSplat & 20.459 & 0.729 & 0.267 & 19.734 & 0.694 & 0.290 & 21.185 & 0.712 & 0.351 & 18.582 & 0.637 & 0.440 \\
        & MVSplat & 20.882 & 0.761 & 0.233 & 19.726 & 0.743 & 0.262 & 14.946 & 0.472 & 0.544 & 13.061 & 0.407 & 0.608 \\
        % & FreeSplat & 20.630 & 0.747 & 0.313 & 21.259 & 0.767 & 0.309 & 28.304 & 0.849 & 0.205 & \textbf{27.040} & 0.832 & 0.224  \\
        \midrule
        \multirow{5}{*}[1em]{\centering \shortstack[l]{\emph{Pose-} \\ \emph{Free}}} 
        % & Splatt3R & - & - & - & - & - & - & - & - & - & - & - & - \\
        & NoPoSplat & 16.299 & 0.552 & 0.397 & 14.372 & 0.469 & 0.520 & 16.940 & 0.560 & 0.425 & 12.939 & 0.397 & 0.588 \\
        & CoPoNeRF & 18.299 & 0.655 & 0.559 & 18.984 & 0.692 & 0.553 & 20.247 & 0.775 & 0.535 & 19.821 & 0.772 & 0.542 \\
        & VicaSplat & 24.537 & 0.814 & 0.162 & 24.502 & 0.806 & 0.164 & 26.673 & 0.856 & 0.188 & 23.656 & 0.777 & 0.262 \\
        & AnySplat & 18.860 & 0.654 & - & 20.273 & 0.602 & - & 21.075 & 0.722 & - & 19.303 & 0.701 & - \\        
        & \textbf{Ours} & \textbf{26.360} & \textbf{0.866} & \textbf{0.129} & \textbf{26.629} & \textbf{0.874} & \textbf{0.118} & \textbf{28.324} & \textbf{0.891} & \textbf{0.161} & \textbf{26.019} & \textbf{0.858} & \textbf{0.193} \\
        \bottomrule
    \end{tabular}
    \end{adjustbox}

    \caption{\textbf{Comparison with 4 and 8-view settings on the RE10k~\cite{RE10K} and ScanNet~\cite{ScanNet} datasets}. }
    % Our method outperforms previous baseline methods on RE10k.}
    % It also exceeds FreeSplat (trained on ScanNet dataset) and exhibits strong generalization capability on ScanNet even not being specifically trained on this dataset.}

\label{tab:re10k_scannet_mv}
\end{table*}

\begin{table}[!t]
    \centering
    \begin{adjustbox}{max width=\linewidth}
    \setlength{\tabcolsep}{0.08cm}
    \renewcommand{\arraystretch}{1.0}
    \begin{tabular}{lcccccc}
        \toprule
        & \multicolumn{3}{c}{2View} & \multicolumn{3}{c}{8View} \\ 
        \cmidrule(lr){2-4} \cmidrule(lr){5-7}
        Method & PSNR$\uparrow$ & SSIM$\uparrow$ & LPIPS$\downarrow$ & PSNR$\uparrow$ & SSIM$\uparrow$ & LPIPS$\downarrow$ \\
        \midrule
        Anysplat~\cite{jiang2025anysplat} & 17.329 & \textbf{0.401} & 0.295 & 18.805 & 0.434 & \textbf{0.308} \\
        \textbf{Ours} & \textbf{17.331} & 0.393 & \textbf{0.283} & \textbf{19.196} & \textbf{0.491} & 0.355 \\
        % \midrule
        % \multirow{2}{*}{\shortstack[l]{\emph{Extrapolation}}} 
        % % & Splatt3R & - & - & - & - & - & - \\
        % & Anysplat~\cite{jiang2025anysplat} & - & - & - & - & - & - \\
        % & \textbf{Ours} & - & - & - & - & - & - \\
        % & \textbf{Ours} & \textbf{25.074} & \underline{0.837} & \textbf{0.158} & \underline{25.766} & \underline{0.778} & \underline{0.192} \\
        \bottomrule
    \end{tabular}
    \end{adjustbox}
\caption{\textbf{Zero-shot generalization on Mip-NeRF360~\cite{barron2022mip} dataset}}
\label{tab:zero_shot_generalization}
\end{table}

\begin{table}[!t]
    \centering
    \begin{adjustbox}{max width=\linewidth}
    \setlength{\tabcolsep}{0.08cm}
    \renewcommand{\arraystretch}{1.0}
    \begin{tabular}{l|cccccc}
        \toprule
        View Nums & PSNR$\uparrow$ & SSIM$\uparrow$ & LPIPS$\downarrow$ & mIoU$\uparrow$ & Acc$\uparrow$ & rel$\downarrow$ \\
        \midrule
        2 Views & 25.27 & 0.867 & 0.141 & 0.5418 & 0.8182 & 4.72 \\
        4 Views & 25.45 & 0.845 & 0.145 & 0.4018 & 0.7616 & 5.76 \\
        8 Views & 24.76 & 0.848 & 0.154 & 0.5539 & 0.8184 & 7.88 \\
        % \textbf{Ours} & - & - & - & - & - & - \\
        \bottomrule
    \end{tabular}
    \end{adjustbox}
\caption{\textbf{Arbitrary View model training and evaluation on the ScanNet~\cite{ScanNet} dataset.}}
\label{tab:arbitrary_view_exp}
\end{table}

\subsection{Experiment Results}

\paragraph{Semantic 3D Reconstruction} 

% As presented in \cref{tab:scannet_semantic} and \cref{fig:scannet_sem}, Uni3R establishes a new state-of-the-art across multiple 3D tasks, and produces remarkably coherent and precise semantic interpretations. Critically, Uni3R does not merely replicate the features of LSeg for alignment, it significantly improves upon it attributed to the robust 3D geometric priors. Furthermore, while methods like LSM necessitate ground-truth point clouds for superivision during training, Uni3R completely obviates the reliance, underscoring its enhanced practicality and scalability for real-world applications.

% As shown in \cref{tab:scannet_semantic} and \cref{fig:scannet_sem}, Uni3R establishes a new state of the art across multiple 3D tasks, producing coherent and precise semantic interpretations. Unlike LSeg, which performs 2D semantic segmentation, Uni3R endows each 3D Gaussian with semantic features, thereby constructing a unified and robust 3D semantic prior that bridges geometry, semantics and rendering. This design not only enhances alignment quality but also substantially improves semantic consistency in 3D space. Furthermore, while methods such as LSM require ground-truth point clouds for supervision, Uni3R eliminates this dependency, demonstrating superior practicality and scalability for real-world applications.

As shown in \cref{tab:scannet_semantic} and \cref{fig:scannet_sem}, Uni3R establishes a new state of the art across multiple 3D tasks, producing coherent and precise semantic interpretations. 
While Uni3R is supervised by LSeg, it outperforms by resolving 2D view-dependent ambiguities through 3D spatial fusion. In \cref{fig:scannet_sem}, LSeg’s 2D predictions are incorrect for the sofa due to local view. Uni3R, however, aggregates features across multiple views into a unified 3D representation. The underlying multi-view geometry acts as a spatial filter that `votes out' inconsistent 2D errors. Thus, Uni3R not only mimics LSeg, but also leverages 3D consistency to produce a denoised, robust semantic prediction.
Furthermore, while methods such as LSM require ground-truth point clouds for supervision, Uni3R eliminates this dependency, demonstrating superior practicality and scalability for real-world applications.

\paragraph{Comparison with Per-Scene Optimized Methods}
To evaluate efficiency and generalization, we compare Uni3R with the per-scene optimized Feature 3DGS~\cite{Feature3DGS}. Such methods rely on Structure-from-Motion~\cite{SfM} to estimate camera poses, leading to high computational overhead and poor scalability. 
In contrast, as shown in \cref{tab:scannet_semantic_multiviews} and \cref{tab:zero_shot_generalization}, Uni3R demonstrates strong generalization by reconstructing consistent 3D geometry, rendering and semantics from unposed multiview inputs. Notably, it achieves superior performance in both novel view synthesis and open-vocabulary segmentation, offering a substantial speed advantage over traditional per-scene optimization methods. 
% while reducing reconstruction time to approximately 0.6 seconds. 
% offering a substantial speed advantage over traditional per-scene optimization methods.

\paragraph{Novel-View Synthesis}
% \subsubsection{Baselines and Metrics}
% \subsubsection{Two View Novel-View Synthesis}

As shown in \cref{tab:re10k_acid_2view}, Uni3R outperforms pose-dependent methods, such as PixelSplat and MVSplat, by a clear margin (1.7dB), and slightly surpasses baseline model NoPoSplat with a gain of 0.2dB on the RE10k dataset.

% As reported in \cref{tab:re10k_acid_2view}, our model outperforms pose-dependent methods such as PixelSplat~\cite{pixelSplat} and MVSplat~\cite{MVSplat} by approximately 1.7dB. It also achieves performance comparable to NoPoSplat~\cite{NoPoSplat}. Additionally, Uni3R demonstrates competitive results, with only a gap of 0.2dB compared to NoPoSplat. Overall, Uni3R achieves performance on par with state-of-the-art approaches on both datasets in terms of novel view synthesis.

% \subsubsection{Multi-View Novel-View Synthesis}

\cref{fig:re10k_sequential} demonstrates Uni3R consistently produces more detailed and structurally coherent constructions. For example, in the Pool Scene, it recovers the forest area with sharper geometry, while VicaSplat shows blurring and discontinuities. These results highlight Uni3R’s ability to preserve fine details and structural consistency in pose-free multi-view 3D reconstruction.

The quantitative results in \cref{tab:re10k_scannet_mv} further validate the effectiveness of Uni3R in aggregating information across multiple views. 
% Uni3R consistently outperforms all baselines under both 4-view and 8-view settings, achieving the highest PSNR. Notably, our method achieves an average improvement of 2.0 dB over VicaSplat~\cite{VicaSplat}, a strong sequential baseline designed for unposed video input, highlighting the superior generalization and multi-view integration capabilities of Uni3R. For NoPoSplat (Ye et al. 2025), we report the results directly from the VicaSplat paper. Furthermore, we test our performance against anysplat number on zero-shot dataset Mip-NeRF360, our model consistently gets better performance.
Uni3R consistently outperforms all baselines under both 4-view and 8-view settings. Notably, it delivers an average improvement of 2.0 dB over VicaSplat~\cite{VicaSplat}, a strong sequential baseline designed for unposed video inputs, demonstrating Uni3R’s superior generalization and multi-view integration capabilities. 
% For NoPoSplat~\cite{NoPoSplat}, we report the results directly from the VicaSplat paper for fair comparison. 
Furthermore, Uni3R surpasses AnySplat~\cite{jiang2025anysplat} on the zero-shot Mip-NeRF360~\cite{barron2022mip} dataset in \cref{tab:zero_shot_generalization} and underscoring its robustness and cross-domain generalization ability.

\subsection{Analysis and Ablations}

\paragraph{Results on Different Input View Numbers}

To demonstrate Uni3R’s ability to handle arbitrary view inputs, we report results in \cref{tab:arbitrary_view_exp}.
% To isolate the effect of input view quantity, we fix the view range and evaluate across varying numbers of input views. As shown in \cref{tab:View_nums}, Uni3R consistently improves in appearance, geometry, and semantic accuracy with more views.
Unlike Vicasplat~\cite{VicaSplat}, which focuses solely on sequential rendering, and LSM~\cite{LSM}, which reconstructs semantic and radiance fields but is restricted to two-view inputs, Uni3R is the first unified model to jointly reconstruct radiance and semantic fields from unposed multiview images. This experiment highlights Uni3R's ability to handle long sequences and wide-baseline configurations, producing high-fidelity and semantically consistent 3D reconstructions in a single feed-forward pass.

% \begin{table}[!t]
%     \centering
%     \footnotesize
%     \begin{adjustbox}{max width=\textwidth}
%     \setlength{\tabcolsep}{0.08cm}
%     \renewcommand{\arraystretch}{1.0}
%     \begin{tabular}{l|ccccccc}
%         \toprule
%         View Nums & mIoU$\uparrow$ & Acc.$\uparrow$ & rel$\downarrow$ & $\tau\uparrow$ & PSNR$\uparrow$ & SSIM$\uparrow$ & LPIPS$\downarrow$\\
%         \midrule
%         % 2 views & - & - & - & - & - & -  & - \\
%         4 views & 0.523 & 0.749 & 9.9 & 31.4 & 16.43 & 0.675 & 0.418 \\
%         8 views & 0.593 & 0.821 & 5.6 & 48.4 & 21.15 & 0.767 & 0.336 \\
%         16 views & 0.629 & 0.834 & 4.7 & 51.4 & 23.36 & 0.808 & 0.294 \\
%         % \textbf{Ours} & \textbf{-} & \textbf{-} & \textbf{-} & \textbf{-} \\
%         \bottomrule
%     \end{tabular}
%     \end{adjustbox}

%     \caption{\textbf{Ablation Study on Input View numbers.} We fix the view range and evaluate the reconstruction quality under varying numbers of input views.}
%     % This experiment highlights Uni3R’s ability to handle long video sequences and wide-baseline scene reconstruction effectively.}
%     % \vspace*{-6mm}
% \label{tab:View_nums}
% \end{table}

\paragraph{Ablation Study on Our Modules}

We conduct ablation studies on Uni3R to analyze the impact of different supervisory signals and architectural components (see \cref{tab:ablation_study}). Removing the semantic loss causes a severe collapse in segmentation accuracy, underscoring its necessity for open-vocabulary semantic learning. Excluding the rendering loss leads to non-convergence, confirming its critical role in guiding 3D reconstruction. When the geometric loss is removed, the model exhibits degraded 3D consistency (higher depth error and lower $\tau$), validating its effectiveness in improving point cloud distribution and depth alignment. The scale-invariant constraint contributes to rendering stability across scenes with varying depth ranges, while the intrinsic embedding improves robustness by aligning scenes of varying scales into a consistent geometric space. Overall, these results demonstrate that Uni3R’s unified supervision of semantic, radiance, and geometric fields is essential for achieving high-fidelity and semantically consistent 3D reconstruction.

% We perform ablations on Uni3R, including semantic, photometric, and geometric supervision (see \cref{tab:ablation_study}). Removing semantic loss leads to a collapse in mIoU, highlighting its necessity for open-vocabulary learning. Without rendering loss, the model fails to converge, confirming its central role in 3D reconstruction guidance. Removing geometry loss degrades 3D consistency, confirming its role in improving point cloud distribution and depth alignment. Moreover, removing the scale-invariant constraint results in rendering quality, showing its stable optimization across scenes with varying depth scales. In general, these results highlight that Uni3R’s unified training of semantic, radiance, and geometric fields is critical to achieving high-fidelity 3D semantic reconstruction.

\begin{table}[!t]
    \centering
    \footnotesize
    \begin{adjustbox}{max width=\textwidth}
    \setlength{\tabcolsep}{0.08cm}
    \renewcommand{\arraystretch}{1.0}
    \begin{tabular}{l|ccccc}
        \toprule
        Method & mIoU$\uparrow$ & rel$\downarrow$ & $\tau\uparrow$ & PSNR$\uparrow$ & SSIM$\uparrow$ \\
        \midrule
        frozen all transfomer layers & 0.0634 & 51.1 & 7.6 & 5.49 & 0.052 \\
        w/o semantic loss & 0.0183 & 5.8 & 47.4  & 25.38 & 0.869 \\
        w/o rendering loss & 0.2653 & N/A & N/A & N/A & N/A \\
        w/o scale invariant & 0.5382 & 5.8 & 47.9 & 24.95 & 0.861  \\
        w/o intrinsic embedding & 0.5471 & 6.5 & 42.5 & 25.35 & 0.871 \\
        w/o geo. loss & \textbf{0.5541} & 5.6 & 48.2 & \textbf{25.53} & 0.872 \\
        \textbf{Ours} & 0.5484 & \textbf{3.9} & \textbf{61.2} & \textbf{25.53} & \textbf{0.873} \\
        \bottomrule
    \end{tabular}
    \end{adjustbox}

    \caption{\textbf{Ablation Study on different modules.} We evaluate the ablated variants of Uni3R, by recording their rendering quality, segmentation performance and geometric accuracy.}

\label{tab:ablation_study}
\end{table}
\section{Conclusion}
\label{sec:conclusion}

Uni3R is a generalizable framework for unified 3D reconstruction and semantic understanding from unposed multi-view images. It predicts a Gaussian-based representation to integrate appearance, geometry, and open-vocabulary semantics in a single forward pass. To address geometric inaccuracies under RGB-only supervision, we introduce a geometry-guided loss to enhance depth consistency. Uni3R takes a significant step toward scalable, multi-view 3D scene understanding for real-world applications, such as autonomous navigation and real-time 3D perception.

% \section*{Acknowledgements}
% This research was supported by the Ministry of Science and ICT (MSIT) of Korea, under the National Research Foundation (NRF) grant (RS-2024-00337548) and a grant of the Korea-US Collaborative Research Fund (KUCRF) funded by the Ministry of Science and ICT and Ministry of Health \& Welfare, Republic of Korea (grant number: RS-2024-00468417). This work was also supported by Institute for Information \& Communications Technology Planning \& Evaluation (IITP) grant funded by the Korea government (MSIT) (No. RS-2024-00457882, AI Research Hub Project, No. RS-2025-25441838, Development of a human foundation model for human-centric universal artificial intelligence and training of personnel, and No. RS-2020-II201361, Artificial Intelligence Graduate School Program (Yonsei University).

{
    \small
    \bibliographystyle{ieeenat_fullname}
    \bibliography{main}

@String(CVPR= {IEEE Conf. Comput. Vis. Pattern Recog.})

@String(ICCV= {Int. Conf. Comput. Vis.})

@String(ECCV= {Eur. Conf. Comput. Vis.})

@String(BMVC= {Brit. Mach. Vis. Conf.})

@String(ICLR = {Int. Conf. Learn. Represent.})

@String(CVPR  = {CVPR})

@String(ICCV  = {ICCV})

@String(ECCV  = {ECCV})

@String(BMVC  =	{BMVC})

@String(ICLR  = {ICLR})

@inproceedings{NeRF,
  author       = {Ben Mildenhall and
                  Pratul P. Srinivasan and
                  Matthew Tancik and
                  Jonathan T. Barron and
                  Ravi Ramamoorthi and
                  Ren Ng},
  title        = {NeRF: Representing Scenes as Neural Radiance Fields for View Synthesis},
  booktitle    = {Computer Vision - {ECCV} 2020 - 16th European Conference, Glasgow,
                  UK, August 23-28, 2020, Proceedings, Part {I}},
  pages        = {405--421},
  year         = {2020},
}

@article{3DGS,
  author       = {Bernhard Kerbl and
                  Georgios Kopanas and
                  Thomas Leimk{\"{u}}hler and
                  George Drettakis},
  title        = {3D Gaussian Splatting for Real-Time Radiance Field Rendering},
  journal      = {{ACM} Trans. Graph.},
  volume       = {42},
  number       = {4},
  pages        = {139:1--139:14},
  year         = {2023},
}

@inproceedings{pixelNeRF,
  author       = {Alex Yu and
                  Vickie Ye and
                  Matthew Tancik and
                  Angjoo Kanazawa},
  title        = {pixelNeRF: Neural Radiance Fields From One or Few Images},
  booktitle    = {{IEEE} Conference on Computer Vision and Pattern Recognition, {CVPR}
                  2021, virtual, June 19-25, 2021},
  pages        = {4578--4587},
  year         = {2021},
}

@inproceedings{pixelSplat,
  author       = {David Charatan and
                  Sizhe Lester Li and
                  Andrea Tagliasacchi and
                  Vincent Sitzmann},
  title        = {PixelSplat: 3D Gaussian Splats from Image Pairs for Scalable Generalizable
                  3D Reconstruction},
  booktitle    = {{IEEE/CVF} Conference on Computer Vision and Pattern Recognition,
                  {CVPR} 2024, Seattle, WA, USA, June 16-22, 2024},
  pages        = {19457--19467},
  year         = {2024},
}

@inproceedings{MVSplat,
  author       = {Yuedong Chen and
                  Haofei Xu and
                  Chuanxia Zheng and
                  Bohan Zhuang and
                  Marc Pollefeys and
                  Andreas Geiger and
                  Tat{-}Jen Cham and
                  Jianfei Cai},
  title        = {MVSplat: Efficient 3D Gaussian Splatting from Sparse Multi-view Images},
  booktitle    = {Computer Vision - {ECCV} 2024 - 18th European Conference, Milan, Italy,
                  September 29-October 4, 2024, Proceedings, Part {XXI}},
  pages        = {370--386},
  year         = {2024},
}

@inproceedings{DepthSplat,
  author       = {Haofei Xu and
                  Songyou Peng and
                  Marc Pollefeys et al.},
  title        = {DepthSplat: Connecting Gaussian Splatting and Depth},
  booktitle    = {{IEEE/CVF} Conference on Computer Vision and Pattern Recognition,
                  {CVPR} 2025, Nashville, TN, USA, June 11-15, 2025},
  pages        = {16453--16463},
  publisher    = {Computer Vision Foundation / {IEEE}},
  year         = {2025},
  url          = {https://openaccess.thecvf.com/content/CVPR2025/html/Xu\_DepthSplat\_Connecting\_Gaussian\_Splatting\_and\_Depth\_CVPR\_2025\_paper.html},
  bibsource    = {dblp computer science bibliography, https://dblp.org}
}

@inproceedings{LangSplat,
  author       = {Minghan Qin and
                  Wanhua Li and
                  Jiawei Zhou and
                  Haoqian Wang and
                  Hanspeter Pfister},
  title        = {LangSplat: 3D Language Gaussian Splatting},
  booktitle    = {{IEEE/CVF} Conference on Computer Vision and Pattern Recognition,
                  {CVPR} 2024, Seattle, WA, USA, June 16-22, 2024},
  pages        = {20051--20060},
  year         = {2024},
}

@inproceedings{Feature3DGS,
  author       = {Shijie Zhou and
                  Haoran Chang and
                  Sicheng Jiang and
                  Zhiwen Fan and
                  Zehao Zhu and
                  Dejia Xu and
                  Pradyumna Chari and
                  Suya You and
                  Zhangyang Wang and
                  Achuta Kadambi},
  title        = {Feature 3DGS: Supercharging 3D Gaussian Splatting to Enable Distilled
                  Feature Fields},
  booktitle    = {{IEEE/CVF} Conference on Computer Vision and Pattern Recognition,
                  {CVPR} 2024, Seattle, WA, USA, June 16-22, 2024},
  pages        = {21676--21685},
  year         = {2024},
}

@inproceedings{LSM,
  author       = {Zhiwen Fan and
                  Jian Zhang and
                  Wenyan Cong and
                  Peihao Wang and
                  Renjie Li and
                  Kairun Wen and
                  Shijie Zhou and
                  Achuta Kadambi and
                  Zhangyang Wang and
                  Danfei Xu and
                  Boris Ivanovic and
                  Marco Pavone},
  title        = {Large Spatial Model: End-to-end Unposed Images to Semantic 3D},
  booktitle    = {Advances in Neural Information Processing Systems 38: Annual Conference
                  on Neural Information Processing Systems 2024, NeurIPS 2024, Vancouver,
                  BC, Canada, December 10 - 15, 2024},
  year         = {2024},
}

@article{UniForward,
  author       = {Qijian Tian and
                  Xin Tan and
                  Jingyu Gong and
                  Yuan Xie and
                  Lizhuang Ma},
  title        = {UniForward: Unified 3D Scene and Semantic Field Reconstruction via
                  Feed-Forward Gaussian Splatting from Only Sparse-View Images},
  journal      = {CoRR},
  volume       = {abs/2506.09378},
  year         = {2025},
  url          = {https://doi.org/10.48550/arXiv.2506.09378},
  doi          = {10.48550/ARXIV.2506.09378},
  eprinttype    = {arXiv},
  eprint       = {2506.09378},
  bibsource    = {dblp computer science bibliography, https://dblp.org}
}

@inproceedings{DUSt3R,
  author       = {Shuzhe Wang and
                  Vincent Leroy and
                  Yohann Cabon and
                  Boris Chidlovskii and
                  J{\'{e}}r{\^{o}}me Revaud},
  title        = {DUSt3R: Geometric 3D Vision Made Easy},
  booktitle    = {{IEEE/CVF} Conference on Computer Vision and Pattern Recognition,
                  {CVPR} 2024, Seattle, WA, USA, June 16-22, 2024},
  pages        = {20697--20709},
  year         = {2024},
}

@inproceedings{VGGT,
  author       = {Jianyuan Wang and
                  Minghao Chen and
                  Nikita Karaev and
                  Andrea Vedaldi and
                  Christian Rupprecht and
                  David Novotn{\'{y}}},
  title        = {{VGGT:} Visual Geometry Grounded Transformer},
  booktitle    = {{IEEE/CVF} Conference on Computer Vision and Pattern Recognition,
                  {CVPR} 2025, Nashville, TN, USA, June 11-15, 2025},
  pages        = {5294--5306},
  year         = {2025},
}

@article{RE10K,
  title={Stereo magnification: Learning view synthesis using multiplane images},
  author={Zhou, Tinghui and Tucker, Richard and Flynn, John and Fyffe, Graham and Snavely, Noah},
  journal={arXiv preprint arXiv:1805.09817},
  year={2018}
}

@inproceedings{ScanNet,
  author       = {Angela Dai and
                  Angel X. Chang and
                  Manolis Savva and
                  Maciej Halber and
                  Thomas A. Funkhouser and
                  Matthias Nie{\ss}ner},
  title        = {ScanNet: Richly-Annotated 3D Reconstructions of Indoor Scenes},
  booktitle    = {2017 {IEEE} Conference on Computer Vision and Pattern Recognition,
                  {CVPR} 2017, Honolulu, HI, USA, July 21-26, 2017},
  pages        = {2432--2443},
  year         = {2017},
}

@inproceedings{ScanNet++,
  author       = {Chandan Yeshwanth and
                  Yueh{-}Cheng Liu and
                  Matthias Nie{\ss}ner and
                  Angela Dai},
  title        = {ScanNet++: {A} High-Fidelity Dataset of 3D Indoor Scenes},
  booktitle    = {{IEEE/CVF} International Conference on Computer Vision, {ICCV} 2023,
                  Paris, France, October 1-6, 2023},
  pages        = {12--22},
  publisher    = {{IEEE}},
  year         = {2023},
  url          = {https://doi.org/10.1109/ICCV51070.2023.00008},
  doi          = {10.1109/ICCV51070.2023.00008},
  bibsource    = {dblp computer science bibliography, https://dblp.org}
}

@inproceedings{SfM,
  author       = {Frank Dellaert and
                  Steven M. Seitz and
                  Charles E. Thorpe and
                  Sebastian Thrun},
  title        = {Structure from Motion without Correspondence},
  booktitle    = {2000 Conference on Computer Vision and Pattern Recognition {(CVPR}
                  2000), 13-15 June 2000, Hilton Head, SC, {USA}},
  pages        = {2557--2564},
  year         = {2000},
}

@inproceedings{MVS,
  author       = {Michael Bleyer and
                  Christoph Rhemann and
                  Carsten Rother},
  title        = {PatchMatch Stereo - Stereo Matching with Slanted Support Windows},
  booktitle    = {British Machine Vision Conference, {BMVC} 2011, Dundee, UK, August
                  29 - September 2, 2011. Proceedings},
  pages        = {1--11},
  year         = {2011},
}

@inproceedings{MVSNeRF,
  author       = {Anpei Chen and
                  Zexiang Xu and
                  Fuqiang Zhao and
                  Xiaoshuai Zhang and
                  Fanbo Xiang and
                  Jingyi Yu and
                  Hao Su},
  title        = {MVSNeRF: Fast Generalizable Radiance Field Reconstruction from Multi-View
                  Stereo},
  booktitle    = {2021 {IEEE/CVF} International Conference on Computer Vision, {ICCV}
                  2021, Montreal, QC, Canada, October 10-17, 2021},
  pages        = {14104--14113},
  year         = {2021},
}

@inproceedings{MASt3R,
  author       = {Vincent Leroy and
                  Yohann Cabon and
                  J{\'{e}}r{\^{o}}me Revaud},
  title        = {Grounding Image Matching in 3D with MASt3R},
  booktitle    = {Computer Vision - {ECCV} 2024 - 18th European Conference, Milan, Italy,
                  September 29-October 4, 2024, Proceedings, Part {LXXII}},
  volume       = {15130},
  pages        = {71--91},
  year         = {2024},
}

@article{Splatt3R,
  author       = {Brandon Smart and
                  Chuanxia Zheng and
                  Iro Laina and
                  Victor Adrian Prisacariu},
  title        = {Splatt3R: Zero-shot Gaussian Splatting from Uncalibrated Image Pairs},
  journal      = {CoRR},
  volume       = {abs/2408.13912},
  year         = {2024},
  url          = {https://doi.org/10.48550/arXiv.2408.13912},
  doi          = {10.48550/ARXIV.2408.13912},
  eprinttype    = {arXiv},
  eprint       = {2408.13912},
  bibsource    = {dblp computer science bibliography, https://dblp.org}
}

@inproceedings{NoPoSplat,
  author       = {Botao Ye and
                  Sifei Liu and
                  Haofei Xu and
                  Xueting Li and
                  Marc Pollefeys and
                  Ming{-}Hsuan Yang and
                  Songyou Peng},
  title        = {No Pose, No Problem: Surprisingly Simple 3D Gaussian Splats from Sparse
                  Unposed Images},
  booktitle    = {The Thirteenth International Conference on Learning Representations,
                  {ICLR} 2025, Singapore, April 24-28, 2025},
  year         = {2025},
}

@inproceedings{CLIP,
  author       = {Alec Radford and
                  Jong Wook Kim and
                  Ilya Sutskever et al.},
  title        = {Learning Transferable Visual Models From Natural Language Supervision},
  booktitle    = {Proceedings of the 38th International Conference on Machine Learning,
                  {ICML} 2021, 18-24 July 2021, Virtual Event},
  pages        = {8748--8763},
  year         = {2021},
}

@inproceedings{DTU,
  author       = {Rasmus Ramsb{\o}l Jensen and
                  Anders Lindbjerg Dahl and
                  George Vogiatzis and
                  Engin Tola and
                  Henrik Aan{\ae}s},
  title        = {Large Scale Multi-view Stereopsis Evaluation},
  booktitle    = {2014 {IEEE} Conference on Computer Vision and Pattern Recognition,
                  {CVPR} 2014, Columbus, OH, USA, June 23-28, 2014},
  pages        = {406--413},
  publisher    = {{IEEE} Computer Society},
  year         = {2014},
  url          = {https://doi.org/10.1109/CVPR.2014.59},
  doi          = {10.1109/CVPR.2014.59},
  bibsource    = {dblp computer science bibliography, https://dblp.org}
}

@inproceedings{ACID,
  author       = {Andrew Liu and
                  Ameesh Makadia and
                  Richard Tucker and
                  Noah Snavely and
                  Varun Jampani and
                  Angjoo Kanazawa},
  title        = {Infinite Nature: Perpetual View Generation of Natural Scenes from
                  a Single Image},
  booktitle    = {2021 {IEEE/CVF} International Conference on Computer Vision, {ICCV}
                  2021, Montreal, QC, Canada, October 10-17, 2021},
  pages        = {14438--14447},
  publisher    = {{IEEE}},
  year         = {2021},
  url          = {https://doi.org/10.1109/ICCV48922.2021.01419},
  doi          = {10.1109/ICCV48922.2021.01419},
  bibsource    = {dblp computer science bibliography, https://dblp.org}
}

@inproceedings{LERF,
  author       = {Justin Kerr and
                  Chung Min Kim and
                  Ken Goldberg and
                  Angjoo Kanazawa and
                  Matthew Tancik},
  title        = {{LERF:} Language Embedded Radiance Fields},
  booktitle    = {{IEEE/CVF} International Conference on Computer Vision, {ICCV} 2023,
                  Paris, France, October 1-6, 2023},
  pages        = {19672--19682},
  year         = {2023},
}

@inproceedings{GaussTR,
  author       = {Haoyi Jiang and
                  Liu Liu and
                  Tianheng Cheng and
                  Xinjie Wang and
                  Tianwei Lin and
                  Zhizhong Su and
                  Wenyu Liu and
                  Xinggang Wang},
  title        = {GaussTR: Foundation Model-Aligned Gaussian Transformer for Self-Supervised
                  3D Spatial Understanding},
  booktitle    = {{IEEE/CVF} Conference on Computer Vision and Pattern Recognition,
                  {CVPR} 2025, Nashville, TN, USA, June 11-15, 2025},
  pages        = {11960--11970},
  year         = {2025},
}

@inproceedings{GD,
  author       = {Seungtae Nam and
                  Xiangyu Sun and
                  Gyeongjin Kang and
                  Younggeun Lee and
                  Seungjun Oh and
                  Eunbyung Park},
  title        = {Generative Densification: Learning to Densify Gaussians for High-Fidelity
                  Generalizable 3D Reconstruction},
  booktitle    = {{IEEE/CVF} Conference on Computer Vision and Pattern Recognition,
                  {CVPR} 2025, Nashville, TN, USA, June 11-15, 2025},
  pages        = {26683--26693},
  publisher    = {Computer Vision Foundation / {IEEE}},
  year         = {2025},
  url          = {https://openaccess.thecvf.com/content/CVPR2025/html/Nam\_Generative\_Densification\_Learning\_to\_Densify\_Gaussians\_for\_High-Fidelity\_Generalizable\_3D\_CVPR\_2025\_paper.html},
  bibsource    = {dblp computer science bibliography, https://dblp.org}
}

@inproceedings{PyTorch,
  author       = {Adam Paszke and
                  Sam Gross and
                  Francisco Massa et al.},
  title        = {PyTorch: An Imperative Style, High-Performance Deep Learning Library},
  booktitle    = {Advances in Neural Information Processing Systems 32: Annual Conference
                  on Neural Information Processing Systems 2019, NeurIPS 2019, December
                  8-14, 2019, Vancouver, BC, Canada},
  pages        = {8024--8035},
  year         = {2019},
}

@article{DINOv2,
  author       = {Maxime Oquab and
                  Timoth{\'{e}}e Darcet and
                  Th{\'{e}}o Moutakanni and
                  Huy V. Vo et al.},
  title        = {DINOv2: Learning Robust Visual Features without Supervision},
  journal      = {Trans. Mach. Learn. Res.},
  year         = {2024},
}

@inproceedings{LPIPS,
  author       = {Richard Zhang and
                  Phillip Isola and
                  Alexei A. Efros and
                  Eli Shechtman and
                  Oliver Wang},
  title        = {The Unreasonable Effectiveness of Deep Features as a Perceptual Metric},
  booktitle    = {2018 {IEEE} Conference on Computer Vision and Pattern Recognition,
                  {CVPR} 2018, Salt Lake City, UT, USA, June 18-22, 2018},
  pages        = {586--595},
  publisher    = {Computer Vision Foundation / {IEEE} Computer Society},
  year         = {2018},
}

@inproceedings{DPT,
  author       = {Ren{\'{e}} Ranftl and
                  Alexey Bochkovskiy and
                  Vladlen Koltun},
  title        = {Vision Transformers for Dense Prediction},
  booktitle    = {2021 {IEEE/CVF} International Conference on Computer Vision, {ICCV}
                  2021, Montreal, QC, Canada, October 10-17, 2021},
  pages        = {12159--12168},
  year         = {2021},
}

@inproceedings{LSeg,
  author       = {Boyi Li and
                  Kilian Q. Weinberger and
                  Serge J. Belongie and
                  Vladlen Koltun and
                  Ren{\'{e}} Ranftl},
  title        = {Language-driven Semantic Segmentation},
  booktitle    = {The Tenth International Conference on Learning Representations, {ICLR}
                  2022, Virtual Event, April 25-29, 2022},
  year         = {2022},
}

@article{VicaSplat,
  author       = {Zhiqi Li and
                  Chengrui Dong and
                  Yiming Chen and
                  Zhangchi Huang and
                  Peidong Liu},
  title        = {VicaSplat: {A} Single Run is All You Need for 3D Gaussian Splatting
                  and Camera Estimation from Unposed Video Frames},
  journal      = {CoRR},
  volume       = {abs/2503.10286},
  year         = {2025},
  url          = {https://doi.org/10.48550/arXiv.2503.10286},
  doi          = {10.48550/ARXIV.2503.10286},
  eprinttype    = {arXiv},
  eprint       = {2503.10286},
  bibsource    = {dblp computer science bibliography, https://dblp.org}
}

@article{Pmloss,
  author       = {Duochao Shi and
                  Weijie Wang and
                  Donny Y. Chen and
                  Zeyu Zhang and
                  Jia{-}Wang Bian and
                  Bohan Zhuang and
                  Chunhua Shen},
  title        = {Revisiting Depth Representations for Feed-Forward 3D Gaussian Splatting},
  journal      = {CoRR},
  volume       = {abs/2506.05327},
  year         = {2025},
  url          = {https://doi.org/10.48550/arXiv.2506.05327},
  doi          = {10.48550/ARXIV.2506.05327},
  eprinttype    = {arXiv},
  eprint       = {2506.05327},
  bibsource    = {dblp computer science bibliography, https://dblp.org}
}

@article{Umeyama,
  author       = {Shinji Umeyama},
  title        = {Least-Squares Estimation of Transformation Parameters Between Two
                  Point Patterns},
  journal      = {{IEEE} Trans. Pattern Anal. Mach. Intell.},
  volume       = {13},
  number       = {4},
  pages        = {376--380},
  year         = {1991},
  url          = {https://doi.org/10.1109/34.88573},
  doi          = {10.1109/34.88573},
  bibsource    = {dblp computer science bibliography, https://dblp.org}
}

@inproceedings{mvsgaussian,
  title={Mvsgaussian: Fast generalizable gaussian splatting reconstruction from multi-view stereo},
  author={Liu, Tianqi and Wang, Guangcong and Hu, Shoukang and Shen, Liao and Ye, Xinyi and Zang, Yuhang and Cao, Zhiguo and Li, Wei and Liu, Ziwei},
  booktitle={European Conference on Computer Vision},
  pages={37--53},
  year={2024},
  organization={Springer}
}

@inproceedings{panogs,
  title={Panogs: Gaussian-based panoptic segmentation for 3d open vocabulary scene understanding},
  author={Zhai, Hongjia and Li, Hai and Li, Zhenzhe and Pan, Xiaokun and He, Yijia and Zhang, Guofeng},
  booktitle={Proceedings of the Computer Vision and Pattern Recognition Conference},
  pages={14114--14124},
  year={2025}
}

@article{supergseg,
  title={Supergseg: Open-vocabulary 3d segmentation with structured super-gaussians},
  author={Liang, Siyun and Wang, Sen and Li, Kunyi and Niemeyer, Michael and Gasperini, Stefano and Navab, Nassir and Tombari, Federico},
  journal={arXiv preprint arXiv:2412.10231},
  year={2024}
}

@article{gsemsplat,
  title={GSemSplat: Generalizable Semantic 3D Gaussian Splatting from Uncalibrated Image Pairs},
  author={Wang, Xingrui and Lan, Cuiling and Zhu, Hanxin and Chen, Zhibo and Lu, Yan},
  journal={arXiv preprint arXiv:2412.16932},
  year={2024}
}

@misc{ilrm3d,
      title={iLRM: An Iterative Large 3D Reconstruction Model}, 
      author={Gyeongjin Kang and Seungtae Nam and Xiangyu Sun and Sameh Khamis and Abdelrahman Mohamed and Eunbyung Park},
      year={2025},
      eprint={2507.23277},
      archivePrefix={arXiv},
      primaryClass={cs.CV},
      url={https://arxiv.org/abs/2507.23277}, 
}

@inproceedings{maskadapter,
  author       = {Yongkang Li and
                  Tianheng Cheng and
                  Bin Feng and
                  Wenyu Liu and
                  Xinggang Wang},
  title        = {Mask-Adapter: The Devil is in the Masks for Open-Vocabulary Segmentation},
  booktitle    = {{IEEE/CVF} Conference on Computer Vision and Pattern Recognition,
                  {CVPR} 2025, Nashville, TN, USA, June 11-15, 2025},
  pages        = {14998--15008},
  publisher    = {Computer Vision Foundation / {IEEE}},
  year         = {2025},
}

@inproceedings{barron2022mip,
  title={Mip-nerf 360: Unbounded anti-aliased neural radiance fields},
  author={Barron, Jonathan T and Mildenhall, Ben and Verbin, Dor and Srinivasan, Pratul P and Hedman, Peter},
  booktitle={Proceedings of the IEEE/CVF conference on computer vision and pattern recognition},
  pages={5470--5479},
  year={2022}
}

@article{jiang2025anysplat,
  title={AnySplat: Feed-forward 3D Gaussian Splatting from Unconstrained Views},
  author={Jiang, Lihan and Mao, Yucheng and Xu, Linning and Lu, Tao and Ren, Kerui and Jin, Yichen and Xu, Xudong and Yu, Mulin and Pang, Jiangmiao and Zhao, Feng and others},
  journal={arXiv preprint arXiv:2505.23716},
  year={2025}
}

@article{hu2025ccgs,
  title={Pointmap association and piecewise-plane constraint for consistent and compact 3d gaussian segmentation field},
  author={Hu, Wenhao and Chai, Wenhao and Hao, Shengyu and Cui, Xiaotong and Wen, Xuexiang and Hwang, Jenq-Neng and Wang, Gaoang},
  journal={arXiv preprint arXiv:2502.16303},
  year={2025}
}

@inproceedings{sun2024f3dgs,
  title={F-3dgs: Factorized coordinates and representations for 3d gaussian splatting},
  author={Sun, Xiangyu and Lee, Joo Chan and Rho, Daniel and Ko, Jong Hwan and Ali, Usman and Park, Eunbyung},
  booktitle={Proceedings of the 32nd ACM International Conference on Multimedia},
  pages={7957--7965},
  year={2024}
}

@inproceedings{lee2024compact3dgs,
  title={Compact 3d gaussian representation for radiance field},
  author={Lee, Joo Chan and Rho, Daniel and Sun, Xiangyu and Ko, Jong Hwan and Park, Eunbyung},
  booktitle={Proceedings of the IEEE/CVF Conference on Computer Vision and Pattern Recognition},
  pages={21719--21728},
  year={2024}
}
}

\clearpage
\setcounter{page}{1}
\maketitlesupplementary

% \section{Rationale}
% \label{sec:rationale}
% % 
% Having the supplementary compiled together with the main paper means that:
% % 
% \begin{itemize}
% \item The supplementary can back-reference sections of the main paper, for example, we can refer to \cref{sec:intro};
% \item The main paper can forward reference sub-sections within the supplementary explicitly (e.g. referring to a particular experiment); 
% \item When submitted to arXiv, the supplementary will already included at the end of the paper.
% \end{itemize}
% % 
% To split the supplementary pages from the main paper, you can use \href{https://support.apple.com/en-ca/guide/preview/prvw11793/mac#:~:text=Delete%20a%20page%20from%20a,or%20choose%20Edit%20%3E%20Delete).}{Preview (on macOS)}, \href{https://www.adobe.com/acrobat/how-to/delete-pages-from-pdf.html#:~:text=Choose%20%E2%80%9CTools%E2%80%9D%20%3E%20%E2%80%9COrganize,or%20pages%20from%20the%20file.}{Adobe Acrobat} (on all OSs), as well as \href{https://superuser.com/questions/517986/is-it-possible-to-delete-some-pages-of-a-pdf-document}{command line tools}.

\section{Appendix}

\begin{table}[t!]
    \centering
    \caption{\textbf{Out-of-distribution performance comparison.} Our method shows superior performance when zero-shot evaluation on DTU and ScanNet++ using the model solely trained on RE10k.}
    \begin{adjustbox}{max width=0.7\textwidth}
    \begin{tabular}{l|cc|cc}
        \toprule
         \multirow{2}{*}{Method} & \multicolumn{2}{c}{DTU} & \multicolumn{2}{c}{ScanNet++} \\
        \cmidrule(lr){2-3} \cmidrule(lr){4-5}
          & PSNR $\uparrow$ & LPIPS $\downarrow$ & PSNR $\uparrow$ & LPIPS $\downarrow$ \\
        \midrule
         pixelSplat & 11.551 & 0.633 & 18.434 & 0.277 \\
         MVSplat    & 13.929 & 0.385 & 17.125 & 0.297 \\
         NoPoSplat  & 17.899 & 0.279 & 22.136 & 0.232 \\
         \midrule
         \textbf{Ours}  & \textbf{18.256} & \textbf{0.266} & \textbf{22.221} & \textbf{0.227} \\
        \bottomrule
    \end{tabular}
    \end{adjustbox}
    \label{tab:zero_shot}
\end{table}

\begin{table}[!t]
    \centering
    \caption{Ablation Study for confidence mask ratio (top-K) on the ScanNet dataset under 2-views setup on source views.}
    \begin{adjustbox}{max width=\linewidth}
    \setlength{\tabcolsep}{0.08cm}
    \renewcommand{\arraystretch}{1.0}

    \begin{tabular}{l|cc|cc|ccc}
        \toprule
        Top-k ratio mask & mIoU$\uparrow$ & Acc.$\uparrow$ & rel$\downarrow$ & $\tau\uparrow$ & PSNR$\uparrow$ & SSIM$\uparrow$ & LPIPS$\downarrow$ \\
        \midrule 
        w/o geo. loss & 53.88 & 82.18 & 5.81 & 47.99 & 24.24 & 0.850 & 0.108 \\
        w/ ratio 100\% & 53.86 & 82.46 & 3.92 & 60.75 & 24.24 & \textbf{0.851} & 0.108 \\
        % w/ geo loss and ratio 100\% and \textcolor{red}{lr 0.01} & 53.49 & 82.61 & \textbf{3.73} & \textbf{63.25} & 24.11 & 0.848 & 0.109 \\
        w/ ratio 90\% & \textbf{54.03} & \textbf{82.55} & \textbf{3.87} & \textbf{61.37} & \textbf{24.35} & \textbf{0.851} & \textbf{0.107} \\ 
        % w/ geo loss and ratio 80\% and lr 0.005 & 53.95 & 82.59 & 4.25 & 58.22 & 24.09 & 0.848 & 0.109 \\
        w/ ratio 70\% & 54.00 & 82.47 & 4.55 & 55.28 & 24.17 & 0.848 & 0.111 \\
        \bottomrule
    \end{tabular}
    \end{adjustbox}
\label{tab:conf_ratio}
\end{table}

% \begin{table}[!t]
%     \centering
%     \caption{Ablation Study for confidence mask ratio (top-K) on the ScanNet dataset under 2-views setup on source views. We report source view performance here.}
%     \begin{adjustbox}{max width=\linewidth}
%     \setlength{\tabcolsep}{0.08cm}
%     \renewcommand{\arraystretch}{1.0}

%     \begin{tabular}{l|ccccccc}
%         \toprule
%         ratio & mIoU$\uparrow$ & Acc.$\uparrow$ & rel$\downarrow$ & $\tau\uparrow$ & PSNR$\uparrow$ & SSIM$\uparrow$ & LPIPS$\downarrow$ \\
%         \midrule 
%         100\% & 53.86 & 82.46 & 3.92 & 60.75 & 24.24 & \textbf{0.851} & 0.108 \\
%         90\% & \textbf{54.03} & \textbf{82.55} & \textbf{3.87} & \textbf{61.37} & \textbf{24.35} & \textbf{0.851} & \textbf{0.107} \\ 
%         80\% & 53.95 & 82.59 & 4.25 & 58.22 & 24.09 & 0.848 & 0.109 \\
%         70\% & 54.00 & 82.47 & 4.55 & 55.28 & 24.17 & 0.848 & 0.111 \\
%         \bottomrule
%     \end{tabular}
%     \end{adjustbox}

% \label{tab:conf_ratio}
% \end{table}

\subsection{Results on the DTU and ScanNet++ dataset}

% To demonstrate our model's strong ability on cross-domain generalization, we train Uni3R on the RE10k~\cite{RE10K} dataset and then do evaluation on the DTU~\cite{DTU} and Scannet++~\cite{ScanNet++} dataset, follow the same setting as NoPoSplat~\cite{NoPoSplat}. As shown in Table.1, our model consistently get better performance than all baseline models. 

To evaluate the cross-domain generalization of Uni3R, we follow NoPoSplat~\cite{NoPoSplat}: training on RE10K~\cite{RE10K} dataset and testing on DTU~\cite{DTU} and ScanNet++\cite{ScanNet++} dataset. As shown in \cref{tab:zero_shot}, Uni3R consistently outperforms all baseline methods on the benchmarks.

\subsection{More Ablation Study on confidence parameter setting in geometry-guided loss}

To validate the effectiveness of our confidence mask in geometry-guided loss, we conduct an ablation study by varying the top-K ratio used for supervision. As shown in Table~\ref{tab:conf_ratio}, applying a 90\% confidence mask yields the best performance in mIoU, depth accuracy, and rendering quality, demonstrating that filtering out low-confidence regions improves overall performance.

\begin{figure}[!t]
  \centering
  \caption{\textbf{Model training w/ and w/o geo. loss on 4 views.}}
  \includegraphics[width=\linewidth]{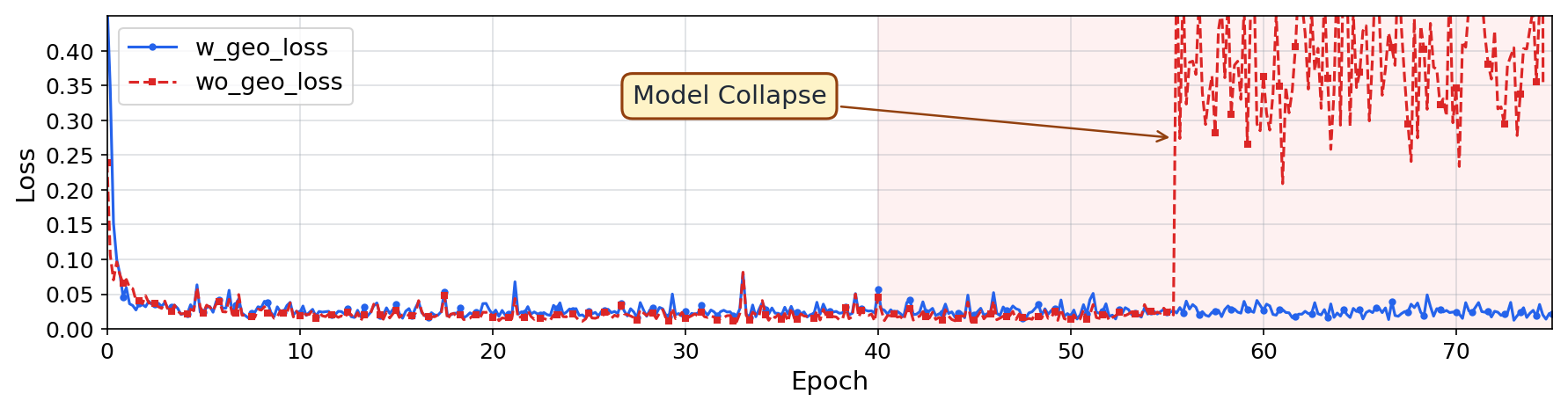}
  \label{fig:uni3r_train}
\end{figure}

Futhermore, the geo. loss from the point map is an essential stability anchor for our unified tasks. In ~\cref{fig:uni3r_train}, training without this constraint under complex setups (e.g., 4-view) leads to model collapse due to the high degree of freedom in Gaussian optimization. Furthermore, ~\cref{tab:conf_ratio} shows in 2-view, the geometry loss significantly improves geometric (47.99 $\rightarrow$ 61.37) while simultaneously improving mIoU (53.88 $\rightarrow$ 54.03). We believe that observing performance improvements across three distinct tasks using only a geometric loss provides a non-trivial insight for the field.

\subsection{Depth Evaluation under Multi-View Settings}

For fair comparison, we follow LSM~\cite{LSM} and adopt Absolute Relative Error (rel) and Inlier Ratio ($\tau$) with a threshold of 1.03 for per-scene depth evaluation. This setting is consistently used throughout the paper.

As shown in \cref{tab:scannet_semantic_multiviews_depth}, Uni3R outperforms the per-scene optimized method on depth estimation under both 8-view and 16-view settings. Notably, our method achieves better depth evaluation performance in one feed-forward.

\begin{table}[t]
    \centering
    \caption{\textbf{Comparison of our method against per-scene optimized methods.}}
    \begin{adjustbox}{max width=\linewidth}
    \setlength{\tabcolsep}{0.08cm}
    \renewcommand{\arraystretch}{1.0}
    \begin{tabular}{l| cc | cc }
        \toprule
        & \multicolumn{2}{c}{8 views} & \multicolumn{2}{c}{16 views} \\ 
        \cmidrule(lr){2-3} \cmidrule(lr){4-5}
        Method & rel$\downarrow$ & $\tau$$\uparrow$ & rel$\downarrow$ & $\tau$$\uparrow$ \\
        \midrule
        Feature-3DGS~\cite{Feature3DGS} & 17.28 & 13.31 & 23.71 & 10.57 \\
        \textbf{Ours} & \textbf{4.46} & \textbf{56.88} & \textbf{5.88} & \textbf{42.88}\\
        \bottomrule
    \end{tabular}
    \end{adjustbox}
\label{tab:scannet_semantic_multiviews_depth}
\end{table}

% \subsection{Model Details}

% More architecture details for our model ...

\subsection{Training and Evaluation Details}

As described in our main paper, we trained our model on three datasets including ScanNet~\cite{ScanNet}, RE10k~\cite{RE10K} and ACID~\cite{ACID}. 

For model training on ACID~\cite{ACID} and RE10K~\cite{RE10K} dataset, we progressively train 2, 4 and 8 view model.
For 2-view training on ACID~\cite{ACID} and RE10K~\cite{RE10K}, we follow NoPoSplat~\cite{NoPoSplat}.
For 4-view training on RE10K, we initialize the model from the 2-view checkpoint and train it on 8×H100 GPUs with a learning rate of 4e-5 for 40,000 iterations, using a batch size of 4 per GPU.
For 8-view training, we further initialize from the 4-view checkpoint and train under the same settings, with a batch size of 1 per GPU. 

For the ScanNet~\cite{ScanNet} dataset, we train Uni3R under 2-view, 8-view, and 16-view settings.
For the 2-view setup, we follow the LSM~\cite{LSM}.
For the 8-view training, we initialize from the 2-view checkpoint and train the model with a learning rate of 5e-5, with a 5 epochs warmup and 50 total epochs. The batch size is set to 4 per GPU.
For the 16-view training, we also initialize from the 2-view checkpoint, with all settings identical to the 8-view setup except for the batch size 2 per GPU.

Additionally, for our arbitrary-view model in the main paper, we uniformly sample 2, 4, and 8 input views from the ScanNet~\cite{ScanNet} dataset and train the model using a batch size of 1 per GPU. The training is performed with a learning rate of 1e-4, including a 10-epoch warm-up and 100 total epochs. As demonstrated in the main paper, our arbitrary-view model achieves consistently comparable performance across different numbers of input views.

\end{document}